\pdfoutput=1

\documentclass[acmsmall]{acmart}

\AtBeginDocument{%
  }

\setcopyright{acmlicensed}
\copyrightyear{2026}
\acmYear{2026}
\acmDOI{XXXXXXX.XXXXXXX}

\acmJournal{TOIS}
\acmVolume{1}
\acmNumber{1}
\acmArticle{1}
\acmMonth{1}

\usepackage{algorithmic}
\usepackage{subcaption}
\usepackage{makecell}
\usepackage{multirow}
\usepackage{tabularx}
\usepackage{siunitx}
\usepackage{alltt}
\usepackage{tcolorbox}
\tcbuselibrary{skins}
\usepackage[colorinlistoftodos,prependcaption,textsize=tiny]{todonotes}
\usepackage{threeparttable}
\usepackage{bm}

\begin{document}

\title[STEP: A Career-Path Recommendation System]{STEP: Career-Path Recommendation via Temporal and Educational Trajectory Modeling}

\author{Iman Johary}
\email{iman.johary@ugent.be}
\affiliation{%
  \institution{AIDA-IDLab, Ghent University}
  \city{Ghent}
  \country{Belgium}
}

\author{Guillaume Bied}
\email{guillaume.bied@ugent.be}
\affiliation{%
  \institution{AIDA-IDLab, Ghent University}
  \city{Ghent}
  \country{Belgium}
}

\author{Alexandru C. Mara}
\email{alexandru.mara@ugent.be}
\affiliation{%
  \institution{AIDA-IDLab, Ghent University}
  \city{Ghent}
  \country{Belgium}
}

\author{Tijl De Bie}
\email{tijl.debie@ugent.be}
\affiliation{%
  \institution{AIDA-IDLab, Ghent University}
  \city{Ghent}
  \country{Belgium}
}

\renewcommand{\shortauthors}{Johary et al.}

\authorsaddresses{%
  \textit{Authors' addresses:}
  Iman Johary, Guillaume Bied, Alexandru C. Mara, and Tijl De Bie,
  AIDA-IDLab, Department of Electronics and Information Systems, Ghent University, Ghent, Belgium.
  \{\href{mailto:iman.johary@ugent.be}{iman.johary}, \href{mailto:guillaume.bied@ugent.be}{guillaume.bied}, \href{mailto:alexandru.mara@ugent.be}{alexandru.mara}, \href{mailto:tijl.debie@ugent.be}{tijl.debie}\}@ugent.be.%
}

\begin{abstract}

  Career paths encode decades of skill acquisition, role transitions, and educational investment, and understanding them at scale underpins workforce planning, labor market policy, and job recommendation.
  Resumes are a rich source of information about career paths: they contain detailed descriptions of work experience, education, and skills.
  Yet their unstructured, heterogeneous, and multilingual nature has long prevented large-scale systematic analysis.
  With the advent of large language models (LLMs), it is now possible to source rich career trajectory data containing temporal and educational signals from unstructured resumes, enabling new opportunities for career-path recommendation.

  Exploiting this opportunity, we present STEP (Sequential Trajectory of Employment Prediction), a novel career-path recommendation system that leverages temporal and educational signals to predict the
  next job in a career trajectory. STEP integrates a time-decay Gated Recurrent Unit (GRU) cell to model temporal dynamics, Feature-wise Linear Modulation (FiLM) conditioned
  on educational attainment, and attention-based sequence pooling to select relevant features for next job prediction.
  To improve internal occupation representation for STEP, we introduce ROUTE, a two-stage contrastive procedure that first adapts a multilingual encoder
  to the career domain via unsupervised denoising autoencoding, then performs supervised contrastive fine-tuning with guided negative selection.

  We evaluate STEP on four datasets of career trajectories, including an improved
  version of our publicly available JobHop dataset, and show that it outperforms state-of-the-art
  baselines in next job prediction.
  The dataset and code are publicly released  to support reproducible career-trajectory research.

\end{abstract}



\maketitle

\clearpage


\begin{figure*}[t!]
  \centering
  \includegraphics[width=\linewidth]{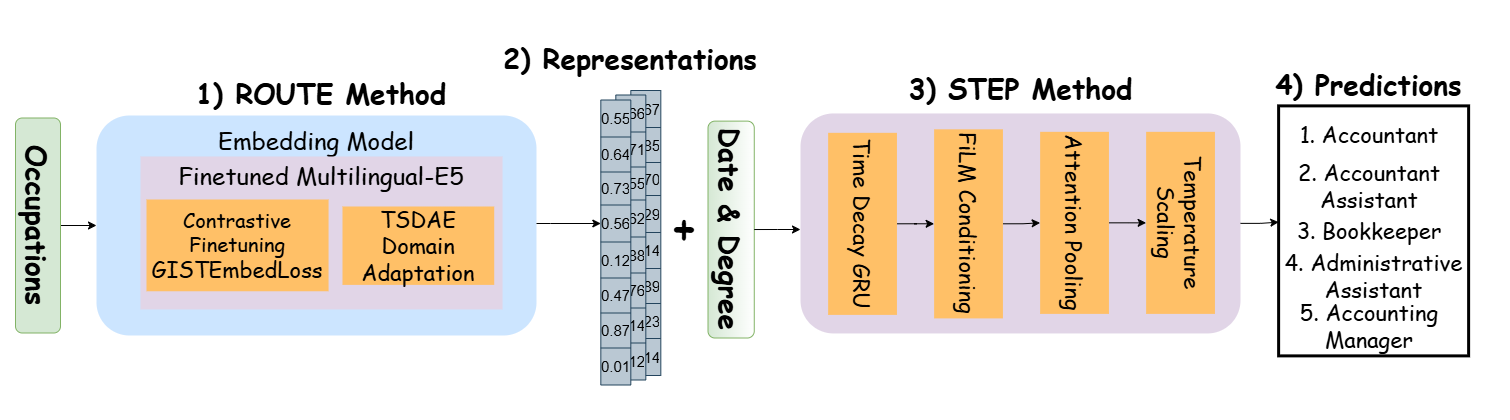}
  \caption{Overview of the career path recommendation pipeline. Each occupation in a career trajectory is encoded by ROUTE (1), an embedding model yielding an occupational representation (2). These embeddings are concatenated with inter-job time intervals and educational attainment (when available) and passed to STEP (3), to produce a ranked list of next job predictions (4). Green boxes denote inputs from the career trajectory datasets; dates and degrees are available only in the JobHop datasets.}
  \label{fig:overview}
\end{figure*}

\section{Introduction}\label{sec:introduction}

Predicting where a career will go next is a problem with concrete stakes: job seekers need guidance on reachable roles,
employers need to identify transferable talent, and policymakers need to anticipate skills gaps before they
emerge~\cite{horton2015labor,alibasic_evaluation_2022}.
These stakes are not hypothetical: randomised controlled trials in public employment services show that guiding job
seekers toward suitable alternative occupations causally improves their labour-market
outcomes~\cite{belot2019advice,belot2025prospects,belot2026automated}. Yet these interventions draw their
recommendations from aggregate occupational transition matrices or skill-based proximity rather than from the
individual's career history; recommendations that exploit a person's full trajectory could target such advice more
precisely, which is the opportunity this work pursues.

Among the available data sources, resumes stand out: unlike job postings, which capture only open roles, or
administrative records, which rarely include job details, they document the full arc of a working life (job titles,
responsibilities, education, and the timing of transitions) in one place.
This richness is buried in unstructured text that spans dozens of formats, multiple languages, and inconsistent
job-title conventions.
Large language models (LLMs) change the calculus: they can normalize job titles to occupational taxonomies, resolve
temporal ambiguity, and extract skill inventories from free-form descriptions at a scale that rule-based and
sequence-labeling pipelines could not approach~\cite{du2024labor,jobhop-v1}, opening large resume corpora to rigorous
career-path modeling for the first time.

Career path recommendation suggests the occupations a worker could plausibly move into next from their
employment history. In its concrete predictive form, career path prediction reduces to next job prediction, the
task of forecasting the next occupation in a career sequence; we use these terms interchangeably.
The literature on this task has advanced considerably, producing methods that span linear transformations
in embedding space~\cite{decorte2023career}, sequential neural architectures~\cite{senger2025methods},
and large language model approaches~\cite{du2024labor}. Progress was nonetheless constrained by the absence of
large-scale, publicly available, richly annotated career trajectory datasets. Most prior work relies on proprietary data
from platforms such as LinkedIn~\cite{li2017nemo} or commercial staffing agencies~\cite{schellingerhout2022},
precluding independent replication. Proprietary datasets derived from large-scale resume corpora~\cite{vafa2022career}
advance the field but remain similarly inaccessible. The few available public benchmarks are either small (Decorte et
al.~\cite{decorte2023career} release $2{,}164$ career histories from anonymized Kaggle resumes,
and OpenResume~\cite{openresume} contains only $301$ real resumes supplemented with synthetic data)
or do not contain additional metadata such as education levels or fine-grained temporal annotations~\cite{karrierewege}.
Recently, however, we introduced JobHop \cite{jobhop-v1} as a large-scale public benchmark,
demonstrating the feasibility of LLM-based extraction from unstructured resumes and establishing a rich baseline with
temporal and educational data.

In this work, we present \textbf{STEP} (Sequential Trajectory of Employment Prediction), a career-path recommendation
model for next job prediction (Figure~\ref{fig:overview}), that takes advantage of temporal and educational signals
from resumes.\footnote{Code for the STEP project---including STEP, ROUTE, and the JobHop~v2 construction pipeline---is publicly available at \url{https://github.com/aida-ugent/Step}.}
STEP is built on the occupational embeddings produced by ROUTE (Representation Of Unique career Trajectories as
Embeddings), a novel domain-adapted occupational embedding procedure developed specifically to improve the representation
quality that STEP relies on for retrieval and prediction.
STEP and ROUTE are trained and evaluated on four career trajectory benchmarks, including \textbf{JobHop~v2},
an improved version of the publicly available JobHop dataset that provides higher-quality extraction, a richer
annotation schema, and finer-grained (five-level) education annotations compared to the original release.

This paper makes the following contributions:
\begin{itemize}
\item \textbf{STEP.}
STEP is a sequential next job prediction model that integrates a time-decay GRU cell conditioning hidden-state
persistence on real-valued inter-job intervals, Feature-wise Linear Modulation (FiLM)~\cite{perez2018film} conditioning on
educational attainment, attention-based sequence pooling selecting most relevant features,
and learnable temperature scaling. STEP achieves state-of-the-art performance on all four primary benchmarks:
Decorte, Karrierewege, and the JobHop and JobHop~v2 datasets.

\item \textbf{ROUTE.} We also improve the occupational embedding quality that STEP relies on by introducing ROUTE,
a two-stage contrastive learning procedure for occupational embeddings: it first adapts a pretrained multilingual encoder
to the career domain via unsupervised denoising autoencoding~\cite{wang2021tsdae}, then performs supervised contrastive
fine-tuning with guided negative selection~\cite{solatorio2024gistembed}. ROUTE is designed to improve the occupational
embedding quality that STEP relies on, and the resulting embeddings improve Recall@10 over a single-stage baseline on all
four primary evaluation datasets, with gains reaching $+4.5$ percentage points on JobHop~v2.

\item \textbf{JobHop~v2.}
JobHop~v2 is an improved version of the publicly available JobHop dataset~\cite{jobhop-v1},
used as one of the four evaluation benchmarks in this work.
Relative to the original release, v2 was extracted using a reasoning controlled LLM leading to higher quality extraction,
zero extraction failures with retry logic, richer extraction schema, and five different education levels.
This enables richer evaluation of both occupational embeddings and next job prediction models.
\end{itemize}

The remainder of this paper is organized as follows.
Section~\ref{sec:related_work} reviews related work. Section~\ref{sec:CPR} presents the STEP career-path recommendation
model including the ROUTE embedding model. Section~\ref{sec:dataset} describes benchmarks and dataset construction, including the improved JobHop~v2.
Section~\ref{sec:evaluation} presents evaluation results, and shows that STEP outperforms a wide range of baselines in
next job prediction. Section~\ref{sec:conclusion} concludes the work. Supplementary technical material---the JobHop~v2
data source and construction pipeline, the extraction evaluation protocol and results, embedding-model exploration and
hyperparameters, STEP architecture details, and additional recommendation results---is collected in the appendices,
which open with a guide to their contents.

\section{Related Work}\label{sec:related_work}

STEP draws on three bodies of prior work, each of which surfaces a specific gap that this paper addresses.
Section~\ref{sec:rw_datasets} surveys existing career trajectory datasets and the information extraction methods that
produce them, motivating the temporally and educationally annotated benchmark that STEP requires.
Section~\ref{sec:rw_representation} discusses occupational representation learning, motivating the domain-adapted
embedding space on which STEP relies for retrieval and prediction.
Section~\ref{sec:rw_prediction} surveys career path prediction (statistical, graph-based, sequential-neural,
and LLM-based) and locates STEP within the broader literature on time-aware sequential recommendation.

\subsection{Career Trajectory Datasets}\label{sec:rw_datasets}

Computational career analysis depends on the quality and scale of available trajectory data.
Proprietary sources such as LinkedIn~\cite{li2017nemo,meng2019hierarchical,zhang2021attentive},
Randstad~\cite{schellingerhout2022}, Zippia~\cite{vafa2022career,du2024labor}, and large in-house Chinese recruitment
corpora~\cite{qin2020tapjfnn,wang2021tactp,zha2024unitrep} provide extensive longitudinal career histories but remain
closed, precluding independent replication. Public survey-based datasets
(NLSY79\footnote{\url{https://www.bls.gov/nls/nlsy79.htm}}/97~\cite{moore2000national},
PSID\footnote{\url{https://psidonline.isr.umich.edu/}}, CPS~\cite{cps2023}) offer nationally representative samples but
are limited in scale or temporal resolution: the CPS, for example, tracks only two consecutive employment spells.
Among recent public datasets, Decorte et al.~\cite{decorte2023career} released $2{,}164$ career histories from anonymized
Kaggle resumes mapped to ESCO~\cite{le2014esco} (the European Union's multilingual classification of occupations and
skills, comprising $3{,}007$ standardized leaf-node occupations organized in a ten-group hierarchy), establishing the
first public benchmark for ESCO-standardized career path prediction. Senger et al.~\cite{karrierewege} constructed the Karrierewege dataset from $568{,}888$ career paths obtained
from the German Federal Employment Agency; occupations were already encoded in the Berufenet taxonomy and subsequently
mapped to ESCO, and free-text job titles and descriptions were synthesized using LLaMA~3.1 to produce richer variants.
Although valuable, Karrierewege does not extract career paths from unstructured resumes: the underlying records are
standardized codes rather than raw text. OpenResume~\cite{openresume} provides a small complementary collection of anonymized
and synthetically augmented resumes that derives from a prior English-language LinkedIn scrape. Large closed corpora such as those underpinning UniTRep~\cite{zha2024unitrep} ($1.19$M
trajectories) and TAPJFNN~\cite{qin2020tapjfnn} ($696$K resumes and $1.49$M applications) show the feasibility of
population-scale representation learning but remain inaccessible to the broader community.
No existing public dataset simultaneously combines large scale, full end-to-end extraction from unstructured text,
rich temporal metadata, education-level annotations, and ESCO standardization.

JobHop~\cite{jobhop-v1} addressed this gap as the first large-scale public dataset of ESCO-coded career trajectories
derived from unstructured resumes. Constructed from $361{,}207$ pseudonymized resumes provided by the Flemish public
employment service, VDAB, the dataset contains $1.67$M work experiences mapped to ESCO occupation codes with quarter-level
temporal annotations and a binary tertiary-degree flag---features that distinguish it from contemporaneous public
benchmarks, which either lack temporal data (Karrierewege) or operate at small scale (Decorte et al.).
The extraction pipeline employs Gemma2-9b-it with one-shot prompting to convert raw pseudonymized text to structured JSON,
validated against $200$ hand-annotated resumes.
It is this extraction pipeline, which transforms raw, pseudonymized resumes into structured ESCO trajectories,
that sets JobHop apart from prior LLM-based extraction efforts.

In contrast to earlier rule-based and neural sequence-labeling extraction
pipelines~\cite{ayishathahira2018}, recent LLM-based approaches have substantially narrowed the gap for
unsupervised occupation extraction. LLM4Jobs~\cite{llm4jobs} and SkillGPT~\cite{skillgpt} combine prompting with
vector-similarity search to extract and standardize occupations and skills without task-specific supervision,
while GoLLIE~\cite{sainz2024gollie} shows that guideline-following instruction tuning substantially improves zero-shot
structured extraction under explicit annotation schemas. Within career-trajectory pipelines,
Decorte et al.~\cite{decorte2023career} used GPT-3.5 to reformat experience sections into a uniform structure,
Senger et al.~\cite{karrierewege} employed LLaMA~3.1 to synthesize free-text descriptions from existing structured data,
and LABOR-LLM~\cite{du2024labor} repurposes general-purpose LLMs as occupational representation models.
The setting addressed in JobHop~\cite{jobhop-v1} is more demanding on several axes: the input corpus consists of
pseudonymized, multilingual (Dutch, French, English) documents containing \texttt{<MASK>} tokens and heterogeneous
formatting, and because privacy constraints preclude cloud inference, the extraction pipeline must run entirely on local
infrastructure---a combination of challenges not addressed by any of the prior methods surveyed above.

\subsection{Occupational Representation Learning}\label{sec:rw_representation}

Early approaches to occupational representation learning relied on graph-based methods (Job2Vec~\cite{zhang2019job2vec},
Dave et al.~\cite{dave2018combined}) or word embeddings computed over job-title text~\cite{kaya2021effectiveness}.
A parallel line of work learned joint representations of job requirements and candidate experiences for person--job fit,
culminating in TAPJFNN~\cite{qin2020tapjfnn}, which combines topic-guided BiLSTM encoders with ability-aware attention
over requirement--experience pairs. Although TAPJFNN targets a recruitment-outcome classification task rather than
taxonomy-anchored retrieval, it establishes the value of attention-based encoding for career text.
More recently, UniTRep~\cite{zha2024unitrep} unifies trajectory-level and market-level representations through a
hypergraph encoder with diachronic node embeddings, linking fine-grained career sequences to coarse-grained labor market
structure.

The field advanced with the maturation of contrastive sentence embedding, rooted in Sentence-BERT~\cite{reimers2019sbert}
and SimCSE~\cite{gao2021simcse}, which established the dual-encoder retrieval paradigm on which modern career encoders
build. CareerBERT~\cite{decorte2023career} adapted this paradigm to the career domain by fine-tuning a sentence
transformer on career--ESCO pairs via contrastive learning; Senger et al.~\cite{senger2025methods} extended the framework
in a reproduction study using
\texttt{all-mpnet-base-v2}~\cite{song2020mpnet}
with multiple negative ranking loss (MNRL)~\cite{henderson2017efficient}. Concurrently, Rosenberger et
al.~\cite{rosenberger2025careerbert} combined domain-adaptive BERT pretraining (jobGBERT~\cite{gnehm2022jobgbert})
with optional TSDAE task-adaptive pretraining~\cite{wang2021tsdae} and SBERT fine-tuning via MNRL on ESCO sentence pairs,
but restricted evaluation to German and frame the task as resume-to-job-advertisement matching rather than career-history
prediction. Concurrent advances in contrastive sentence embedding, including TSDAE~\cite{wang2021tsdae} for unsupervised
domain adaptation, GISTEmbedLoss~\cite{solatorio2024gistembed} for guided negative selection,
Matryoshka representation learning~\cite{kusupati2022matryoshka} for nested embedding dimensions,
and multilingual encoders such as
\texttt{multilingual-E5}~\cite{wang2024multilingual},
offer complementary tools that have not previously been combined for career representation.
No prior work combined multilingual domain-adaptive pretraining with guided contrastive training over career-history
subsequences anchored to ESCO.

\subsection{Career Path Prediction}\label{sec:rw_prediction}

Career path prediction has been pursued through diverse paradigms. Statistical and graph-based methods model
skill--occupation relationships and transition frequencies~\cite{gugnani2018implicit,chang2019bayesian}.
Sequential neural architectures, including LSTM-based models~\cite{li2017nemo,cerilla2023career,schellingerhout2022},
hierarchical career-path-aware networks~\cite{meng2019hierarchical}, and attentive heterogeneous graph embeddings for job
mobility prediction~\cite{zhang2021attentive}, capture temporal dependencies in occupational sequences by conditioning on
profile context or graph-structured neighbourhoods. Transformer-based models pretrained on proprietary resume corpora,
notably CAREER~\cite{vafa2022career}, demonstrate the value of scale but are not publicly reproducible.
Their successor LABOR-LLM~\cite{du2024labor} repurposes general-purpose LLMs for the same task.
Senger et al.~\cite{senger2025methods}, however, report that LLMs generally underperform MLP and LSTM baselines in the
standardized ESCO-based prediction setting, suggesting that next-token generation is misaligned with the retrieval-based
formulation. Recent work pushes sequential modeling in new directions: NAOMI~\cite{yamashita2022looking} forecasts entire
career pathways rather than a single next move, and CAPER~\cite{lee2025caper} integrates continual learning to accommodate
evolving occupation taxonomies.
UniTRep~\cite{zha2024unitrep}, discussed above, is the most directly comparable prior work,
reporting Career Mobility Recommendation results over $800$ companies on a private $1.19$M-trajectory corpus.

Career path prediction in the ESCO setting is an instance of \emph{time-aware sequential recommendation},
a well-studied problem outside the career domain. GRU4Rec~\cite{hidasi2016gru4rec} introduced RNN-based session
recommendation; SASRec~\cite{kang2018sasrec} and BERT4Rec~\cite{sun2019bert4rec} established self-attentive and
bidirectional transformer architectures as strong baselines. Subsequent work has incorporated absolute and relative time
signals: TiSASRec~\cite{li2020tisasrec} encodes pairwise time intervals into self-attention,
and T-LSTM~\cite{baytas2017tlstm} decomposes cell memory by elapsed time for irregular clinical sequences.
Within the career domain, TACTP~\cite{wang2021tactp} applied variable-interval temporal encoding to career sequences in a
collaborative filtering framework on proprietary data, demonstrating the value of time-aware modeling for occupational
transitions.

A key gap in the ESCO-based prediction setting nonetheless remains: the joint modeling of real-valued temporal intervals
and educational attainment within a retrieval-based sequential architecture trained on publicly available data.
STEP addresses this gap by combining a time-decay recurrent cell, FiLM-based degree conditioning,
and attention-based sequence pooling over domain-adapted ESCO occupational embeddings.

\section{Methods}\label{sec:CPR}

This section presents the methods underlying STEP.
Section~\ref{sec:cpr_problem} formalizes next job prediction as a ranking task, independently of any particular
model, and fixes the notation used throughout.
Section~\ref{sec:representation} describes ROUTE, the two-stage representation learning procedure used to produce the
domain-adapted occupational embeddings on which STEP operates. Section~\ref{sec:pred_proposed} introduces STEP itself,
a sequential prediction model that integrates a time-decay recurrent cell, FiLM-based degree conditioning,
attention-based sequence pooling, and learnable temperature scaling into a single architecture trained end-to-end against
the candidate label space.

We first fix terminology used throughout the paper. A \emph{job} $j$ is an instance of a position an
individual actually held, described in free text (a role title and a description); the set of all possible jobs
is denoted $\mathcal{J}$. An \emph{occupation} $o$ is the standardized counterpart of a job,
obtained by mapping jobs to a fixed taxonomy, and the finite set of all occupations in that taxonomy is denoted
$\mathcal{O}$. We identify each occupation with its canonical textual description, itself a prototypical job, so that
$\mathcal{O}$ is a subset of the job set, $\mathcal{O} \subseteq \mathcal{J}$. Since every occupation is
therefore also a job, we state the problem in terms of jobs throughout; wherever the data provides standardized
occupations, jobs can simply be replaced by occupations. In this work, $\mathcal{O}$ is the ESCO taxonomy, 
comprising $K = 3{,}007$ occupations.

\subsection{Problem Formulation}\label{sec:cpr_problem}

Career-path recommendation is formalized as a many-to-one sequence prediction task, which we refer to as next job prediction.
Given an observed career trajectory $(j_1, \ldots, j_{t-1})$, a chronologically ordered sequence of jobs,
the goal is to predict the individual's next job $j_t$ from a finite candidate set
$\mathcal{J}_c \subseteq \mathcal{J}$. The candidate set is dataset-dependent: for datasets standardized to an
occupation taxonomy, $\mathcal{J}_c$ is the set of occupations in the taxonomy, $\mathcal{J}_c = \mathcal{O}$; for datasets of free-text
jobs, $\mathcal{J}_c$ is the set of unique jobs observed in the dataset. As the correct successor
cannot in general be determined with certainty, the prediction is expressed as a ranking: the candidates in
$\mathcal{J}_c$ are ordered by their estimated likelihood of being $j_t$, and the highest-ranked candidates are returned
as recommendations. A prediction is considered better when the ground-truth next job $j_t$ appears higher in the ranking.

Beyond the sequence of positions, a trajectory may carry auxiliary signals.
The inter-job time intervals $\boldsymbol{\delta} = (\delta_2, \ldots, \delta_{t-1})$ record the elapsed time in years
between consecutive positions, where $\delta_s$ is the gap between $j_{s-1}$ and $j_s$. An education-level indicator
$e \in \mathcal{E}$ encodes the highest attained education level, with $\mathcal{E}$ a small set of levels (e.g.,
Secondary, Bachelor, Master, PhD); other signals may be available depending on the data source.
Such signals are not present uniformly across trajectories or datasets, and a method may use or ignore them.

We assume access to a dataset of career trajectories $\mathcal{C} = \{c_1, \ldots, c_N\}$,
where each $c_i = (j_{i,1}, \ldots, j_{i,T_i})$ is a single individual's trajectory and each element $j_{i,t}$ is a
textual representation of a job: either a free-text job (role title and description) formatted as
\texttt{role: \{title\}$\backslash$n description: \{text\}}, or a
standardized occupation description formatted as
\texttt{occupation: \{title\}$\backslash$n description: \{text\}}.
The remainder of this section describes how trajectories and occupations are represented and how the two are compared to
rank candidates and produce recommendations.

\subsection{Career Representation Learning}\label{sec:representation}

This subsection introduces ROUTE, the representation learning procedure that produces the embedding space STEP operates on.

To learn high-quality representations, we cast next job prediction as retrieval in a shared embedding space.
A single encoder $f_\theta: \mathcal{J}^{+} \to \mathbb{R}^d$ maps any free-text input, whether a single job or a career
history rendered as the concatenation of its jobs, to a $d$-dimensional vector, where
$\mathcal{J}^{+} = \bigcup_{n\geq 1}\mathcal{J}^{n}$ is the set of finite job sequences and a candidate job is the
length-one case $\mathcal{J} = \mathcal{J}^{1}$. The same encoder embeds both a candidate job and a career history,
and a history is scored against the embeddings of all candidate jobs in $\mathcal{J}_c$ by vector similarity,
returning the most similar jobs. Under this framing the embedding space is the central object: retrieval quality depends
directly on whether semantically related jobs lie close together, and whether a career history embeds close to its
plausible successor.

This places a heavy burden on the space. Candidate jobs number in the thousands, many are rare or unseen during training,
and their textual descriptions vary widely in surface form, so ranking must be driven by meaning rather than surface
overlap. A general-purpose sentence encoder, untuned to career text, is unlikely to provide such a space; ROUTE therefore
adapts and fine-tunes the encoder to the career domain in two stages. The objective is taken directly from the prediction
task of Section~\ref{sec:cpr_problem}: at every point in a trajectory, the prefix of job experiences observed so far
should embed close to the job that actually comes next. The remainder of this subsection first describes how each
trajectory is decomposed into a set of (history, next job) training examples that make this objective concrete, then
presents the two-stage training procedure that operates on them.

\subsubsection{Training Data Construction}

From each trajectory we construct a set of anchor--positive pairs following the
subsequence decomposition of Decorte et al.~\cite{decorte2023career} (their
\emph{Last} strategy, also adopted by Senger et al.~\cite{senger2025methods}).
For an individual with trajectory $(j_1, j_2, \ldots, j_T)$, every prefix is
paired with the job that immediately follows it:
\begin{equation}
\label{eq:pair_generation}
    \{(\text{concat}(j_1, \ldots, j_{t-1}),\; j_t) \mid t = 2, 3, \ldots, T\},
\end{equation}
where $\text{concat}(\cdot)$ concatenates job experiences with \texttt{<SEP>}
delimiters, so that $\text{concat}(j_1, \ldots, j_{t-1}) \in \mathcal{J}^{+}$ and both the anchor history and the
positive $j_t$ are encoded by the same $f_\theta$. The anchor encodes the career history available at step $t$; the
positive $j_t$ is the next job the model must rank as highly as possible at inference.

This construction yields $T-1$ pairs per trajectory of length $T$, one per
prediction step. Beyond data augmentation, exposing the encoder to prefixes of
every length ensures it produces a usable history representation independent of whether a person
has held two positions or ten, and training each prefix to embed close to its true
next job experience is what shapes the embedding space the retrieval step relies on.

\subsubsection{Two-Stage Domain-Adapted Training}

Two limitations of standard contrastive sentence embedding training on career data motivate the two-stage design.
First, the pretrained encoder has limited exposure to career-domain vocabulary and to the compositional semantics of job
descriptions, yielding suboptimal initial representations for the retrieval task STEP relies on.
Second, random batch composition produces uninformative in-batch negatives: occupations within the same ESCO group share
highly similar descriptions, so randomly sampled negatives often act as false negatives and corrupt the contrastive
signal. ROUTE addresses these limitations through unsupervised domain adaptation (Stage~1,
TSDAE) followed by guided contrastive fine-tuning (Stage~2, GISTEmbedLoss), yielding cleaner,
more semantically separated occupational representations for STEP to operate on.

\paragraph{TSDAE domain adaptation.}
Stage 1 adapts the pretrained encoder to the distributional characteristics of career text via the Transformer-based
Sequential Denoising Auto-Encoder (TSDAE)~\cite{wang2021tsdae}. TSDAE corrupts input sentences by randomly deleting
tokens, encodes the corrupted input, and trains a decoder to reconstruct the original. This denoising objective encourages
the encoder to capture the full semantic content of career-domain text even from partial or noisy inputs,
a property especially relevant given the heterogeneous quality of resume-derived descriptions.
The TSDAE corpus pools all unique text spans from the training data in addition to the ESCO occupation descriptions, 
ensuring that the encoder is exposed to the full range of career-relevant vocabulary.
After TSDAE training, the decoder is discarded and the adapted encoder is retained for Stage 2.

\paragraph{Guided contrastive fine-tuning.}
Stage 2 performs supervised contrastive fine-tuning on the career trajectory pairs with GISTEmbedLoss (Guided In-Sample
Triplet Selection for Embedding Loss)~\cite{solatorio2024gistembed}. GISTEmbedLoss augments standard in-batch contrastive
learning with a frozen \emph{guide model} that provides auxiliary similarity scores. For each batch,
the guide model (\texttt{all-MiniLM-L6-v2}, 384-dimensional, frozen throughout) computes pairwise similarities among all
in-batch items. These scores filter false negatives (semantically similar pairs mistakenly treated as negatives under
random batch composition) and emphasize informative hard negatives, yielding a cleaner gradient signal than standard MNRL.

\paragraph{Base model selection.}
From a systematic comparison of five candidate encoders ({\small\texttt{multilingual-e5- base}}~\cite{wang2024multilingual}, {\small\texttt{multilingual-e5-large}}~\cite{wang2024multilingual},
{\small\texttt{bge-base-en-v1.5}}~\cite{xiao2024cpack}, {\small\texttt{bge-large-en-v1.5}}~\cite{xiao2024cpack},
and {\small\texttt{all- mpnet-base-v2}}~\cite{song2020mpnet}), we select \texttt{multilingual-e5-base} for ROUTE.
It achieves the highest mean reciprocal rank (MRR) among base-scale encoders while requiring substantially less GPU memory and training time than
the 1024-dimensional large variants, which show no consistent improvement despite their increased capacity.
The multilingual pretraining of E5 is also well suited to the Dutch/French/English JobHop corpus.
Full exploration details, including 13 embedding training configurations, appear in
Appendix~\ref{app:embedding_exploration}, and the final fine-tuning hyperparameters for ROUTE are reported in
Appendix~\ref{app:embedding_hyperparams}.

\begin{figure*}[t]
  \centering
  \includegraphics[width=\linewidth,height=0.45\textheight,keepaspectratio]{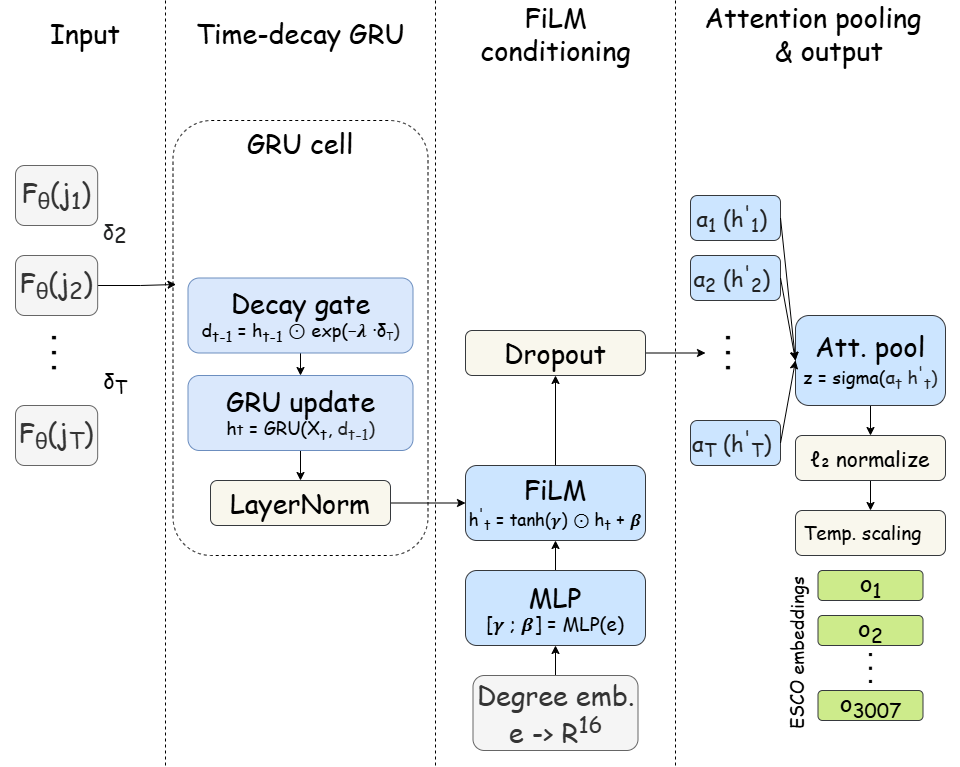}
  \caption{Architecture of STEP (Sequential Trajectory of Employment Prediction). The model processes input embeddings through a time-decay GRU cell conditioned on temporal intervals ($\delta$), followed by layer normalization, FiLM-based degree conditioning, and dropout. Attention pooling aggregates the sequence into a fixed-size representation, which is projected and L2-normalized before computing temperature-scaled cosine similarity against the candidate embeddings (the ESCO occupations for taxonomy-standardized datasets).}
  \label{fig:step_architecture}
\end{figure*}

\subsection{STEP: Sequential Trajectory of Employment Prediction}\label{sec:pred_proposed}

With the ROUTE-trained encoder providing occupational embeddings, we now turn to STEP itself.
STEP is a sequential next job prediction model that integrates four architectural components:
a time-decay GRU cell that conditions hidden-state persistence on inter-job intervals, FiLM-based conditioning on
educational attainment, attention-based sequence pooling, and learnable temperature scaling over the candidate set $\mathcal{J}_c$
(Figure~\ref{fig:step_architecture}).

Formally, let $f_\theta(j_t)$ be the embedding of job $j_t$ produced by the ROUTE-trained
encoder $f_\theta$ introduced in Section~\ref{sec:representation}, which remains frozen during training of STEP.
STEP is a parametric function $g_\phi$ that maps the encoded
trajectory, its inter-job time intervals, and the education level to a single output embedding;
at inference, the full observed trajectory is used as input:
\begin{equation}
\label{eq:cpr_model}
    \hat{\mathbf{y}} = g_\phi\bigl((f_\theta(j_1), \ldots, f_\theta(j_T)),\; (\delta_2, \ldots, \delta_T),\; e\bigr) \in \mathbb{R}^d.
\end{equation}
The predicted embedding $\hat{\mathbf{y}}$ is matched against the candidate embeddings to produce the ranked
recommendations, as detailed below.

\subsubsection{Time-Decay GRU Cell}\label{sec:pred_time_gru}

Standard recurrent cells assume uniform time steps between observations, an assumption strongly violated by career
trajectories, in which individuals may hold positions for months or decades. Prior work on time-aware sequence modeling
outside the career domain, notably the time-aware LSTM of Baytas et al.~\cite{baytas2017tlstm},
which decomposes cell memory by elapsed time for irregularly sampled clinical trajectories,
has demonstrated the value of explicit time-decay on hidden state. Within the career domain,
TACTP~\cite{wang2021tactp} addresses variable-interval sequences through a temporal encoding mechanism in a collaborative
filtering setting. We adopt a complementary approach that inherits T-LSTM's exponential hidden-state decay but applies it
within a retrieval-based architecture over ROUTE-trained ESCO embeddings. Specifically, we introduce an exponential decay
mechanism into the GRU cell~\cite{cho2014learning}.

Before each GRU update, the hidden state is multiplied element-wise by a time-dependent decay factor:
\begin{equation}
\label{eq:time_decay}
    \mathbf{d}_{t-1} = \mathbf{h}_{t-1} \odot \exp(-\boldsymbol{\lambda} \cdot \delta_t),
\end{equation}
where $\boldsymbol{\lambda} \in \mathbb{R}^H$ is a vector of learnable per-dimension decay rates,
parameterized as $\boldsymbol{\lambda} = \text{softplus}(\boldsymbol{\ell})$ with $\boldsymbol{\ell}$ initialized at zero
(no decay at the start of training). The decayed hidden state is then passed to a standard GRU update:
$\mathbf{h}_t = \text{GRU}(\mathbf{x}_t, \mathbf{d}_{t-1})$.

The per-dimension parameterization allows the model to learn that some aspects of career state (e.g.,
domain expertise) decay slowly, while others (e.g., recency of specific technical skills) decay more rapidly.
This formulation is inspired by the continuous-time dynamics of Neural ODEs~\cite{chen2018neural} but avoids the
computational cost of numerical ODE solvers through an analytically tractable exponential decay.
On datasets without temporal information, $\delta_t$ is set to a constant (1.0~year) at inference,
reducing the cell to a standard GRU with a fixed per-dimension scaling.

We choose the GRU over the LSTM~\cite{hochreiter1997long} because career histories in our datasets are short (typically
fewer than 10 positions). The LSTM's dedicated long-term memory cell offers marginal benefit for short sequences while
adding parameters and training complexity; the GRU's simpler gating is sufficient for the temporal dependencies in career
trajectory data.

\subsubsection{FiLM Degree Conditioning}\label{sec:pred_film}

Educational attainment shapes career trajectories in ways that occupational history alone does not fully capture:
a software engineer with a PhD may follow systematically different progressions than one with a bachelor's degree,
even given identical employment records. Rather than concatenating degree information as an additional input feature
(which would conflate it with the occupational signal), we adopt Feature-wise Linear Modulation
(FiLM)~\cite{perez2018film}, conditioning the hidden representation via learned affine transformations:
\begin{equation}
\label{eq:film}
    \mathbf{h}_t' = \tanh(\boldsymbol{\gamma}) \odot \mathbf{h}_t + \boldsymbol{\beta},
    \quad [\boldsymbol{\gamma};\, \boldsymbol{\beta}] = \text{MLP}(\text{Embed}(e)),
\end{equation}
where $e$ is encoded to $\mathbb{R}^{16}$ via a learned embedding, projected to
$(\boldsymbol{\gamma}, \boldsymbol{\beta})$ via a two-layer network, and applied at every time step.
The $\tanh$ on $\boldsymbol{\gamma}$ bounds multiplicative modulation to $[-1, 1]$, preventing training instabilities.
On datasets without degree information, a dedicated ``None'' class in $\mathcal{E}$ is used for every trajectory,
producing a fixed learned affine transformation.

\subsubsection{Attention-Based Sequence Pooling and Temperature Scaling}\label{sec:pred_attention_temp}

\textbf{Attention pooling.}
To aggregate the sequence of degree-conditioned hidden states $(\mathbf{h}_1', \ldots, \mathbf{h}_T')$ into a single
representation, we use learnable attention pooling:
\begin{equation}
\label{eq:attention}
    \alpha_t = \frac{\exp(a(\mathbf{h}_t'))}{\sum_{s=1}^T \exp(a(\mathbf{h}_s'))}, \qquad
    \mathbf{z} = \sum_{t=1}^T \alpha_t\, \mathbf{h}_t',
\end{equation}
where $a(\cdot)$ is a two-layer feed-forward scoring network with $\tanh$ activation. Padding positions are masked with
$-\infty$ before the softmax. This mechanism lets the model weight career steps by predictive relevance rather than rely
solely on the most recent position.

\textbf{Temperature scaling.}
The output embedding $\hat{\mathbf{y}} = \text{normalize}(\mathbf{W}_\text{out}\mathbf{z})$ is $\ell_2$-normalized and
compared against the $\ell_2$-normalized candidate embeddings via cosine similarity.
Each of the $K = |\mathcal{J}_c|$ candidates $j_k \in \mathcal{J}_c$ is embedded by the ROUTE encoder as
$f_\theta(j_k)$; for taxonomy-standardized datasets these are the $K = 3{,}007$ ESCO occupations.
A learnable temperature parameter $\tau$ scales the resulting logits:
\begin{equation}
\label{eq:temperature}
    \ell_k = \frac{\hat{\mathbf{y}}^\top f_\theta(j_k)}{\tau}, \qquad
    \tau = \min\!\bigl(\max(\exp(\log \tau_0),\, 10^{-3}),\; 1\bigr),
\end{equation}
where $\log \tau_0$ is the learnable parameter and $\tau_0$ is initialized at 0.07. Temperature scaling controls the sharpness of the similarity distribution over the large
candidate set ($K = |\mathcal{J}_c|$): lower $\tau$ produces more peaked distributions that assign higher confidence to top
candidates. As shown by the ablation study (Section~\ref{sec:pred_ablation}), this component has the single largest impact
on model performance.

\subsubsection{Training Objective and Optimization}\label{sec:pred_training}

The model is trained end-to-end with cross-entropy over temperature-scaled similarity logits,
with label smoothing ($\epsilon = 0.1$) to mitigate overconfidence:
\begin{equation}
\label{eq:loss}
    \mathcal{L} = -\sum_{k=1}^{K} \tilde{y}_k \log \frac{\exp(\ell_k)}{\sum_{k'=1}^K \exp(\ell_{k'})},
    \qquad
    \tilde{y}_k =
    \begin{cases}
        (1-\epsilon) + \dfrac{\epsilon}{K} & \text{if } k = k^*,\\[4pt]
        \dfrac{\epsilon}{K} & \text{otherwise,}
    \end{cases}
\end{equation}
where $k^*$ is the index of the ground-truth candidate in $\mathcal{J}_c$.
We optimize with AdamW~\cite{loshchilov2019decoupled} (learning rate $10^{-3}$, weight decay $10^{-4}$,
$\beta_1{=}0.9$, $\beta_2{=}0.999$), cosine annealing~\cite{loshchilov2017sgml}, and gradient clipping at $1.0$.
The model is lightweight: with a hidden dimension of 64 and a degree embedding dimension of 16,
it comprises approximately $222{,}000$ trainable parameters (full breakdown in Appendix~\ref{app:model_architecture})
and adds negligible overhead beyond the embedding computation.

\section{Evaluation Datasets}\label{sec:dataset}

This section describes the datasets used to evaluate STEP and ROUTE.
Section~\ref{sec:benchmark_data} surveys the public career trajectory benchmarks used as primary evaluation data,
and Section~\ref{sec:JobHop_v2_dataset} describes JobHop~v2, an improved version of the JobHop dataset released with this
work.

\subsection{Existing Benchmark Datasets}\label{sec:benchmark_data}

In this section, we describe the available benchmark datasets for ESCO-based career path prediction,
which serve as the basis for our evaluation of STEP and ROUTE.

\paragraph{Decorte.}
Decorte et al.~\cite{decorte2023career} introduced the first public benchmark for ESCO-standardized career path
prediction. The dataset comprises $2{,}164$ anonymized career histories derived from a Kaggle resume corpus
(livecareer.com): of $2{,}482$ source resumes, $2{,}473$ were successfully parsed, and after retaining only profiles
with two or more job experiences, $2{,}164$ career histories remain, split into $1{,}720$ training, $217$ validation,
and $227$ test careers (an $80/10/10$ split). Each job experience is mapped to an ESCO v1.1.2 occupation code via a proprietary classifier
(estimated label error rate: ${\sim}2.2\%$), and experience descriptions were reformatted using GPT-3.5 for consistency.
The dataset contains no temporal annotations (no job dates) and no education level information,
and is skewed toward white-collar occupations (predominantly ESCO Sectors~1 and~2). It is publicly available on
HuggingFace (\texttt{TechWolf/anonymous-working-histories}). Alongside the main Decorte dataset we consider a derivative
variant, Decorte-ESCO, in which free-text job titles are replaced by their corresponding ESCO taxonomy descriptions;
this variant probes robustness to standardized rather than resume-language inputs.

\paragraph{Karrierewege.}
Senger et al.~\cite{karrierewege} introduced KARRIEREWEGE, a large-scale career path dataset sourced from the German
Federal Employment Agency (Bundesagentur für Arbeit), comprising $568{,}888$ anonymized career paths.
Occupation titles from the Berufenet classification system were mapped to ESCO v1.2.0 codes via a GPT-4o-mini pipeline
using ISCO-08 codes as a pre-filter, achieving a mapping accuracy of $60.4\%$. Compared to Decorte,
Karrierewege covers a broader and more realistic occupational distribution, with high representation of elementary,
service, and trade occupations (ESCO Sectors~5, 8, and~9). Like Decorte, it contains no temporal or degree annotations.
It is distributed with an $80/10/10$ train/validation/test split over career paths.
Two synthesized free-text variants (Karrierewege-Occ and Karrierewege-CP) were derived from a 100K subset using
Llama~3.1~8B~\cite{senger2025methods}. In the Occ variant, titles and descriptions are generated independently per ESCO
occupation; in the CP variant, the entire career path is paraphrased at once, producing more contextually coherent titles.
These variants enable robustness testing across different text representation styles.

\paragraph{JobHop.}
JobHop~\cite{jobhop-v1} is a large-scale public dataset of real-world career trajectories derived from $361{,}207$
pseudonymized resumes provided by VDAB, the Flemish Public Employment Service. It contains $1.67$~million work
experiences, each mapped to an ESCO occupation code through a two-stage pipeline: LLM-based structured extraction from raw
resume text (Gemma2-9b achieving $82.1\%$ accuracy on a hand-annotated validation set) followed by embedding-based ESCO
normalization ($84.7\%$ ESCO group-level accuracy). A distinguishing feature of JobHop relative to other public benchmarks
is the inclusion of quarter-level temporal annotations alongside a binary tertiary-degree flag,
making it the only large-scale public dataset to combine ESCO-coded occupational transitions with job timing information.
The source resumes are predominantly Dutch ($91.4\%$), reflecting the Flemish labor market,
with smaller portions in French and English. The dataset is publicly available on HuggingFace
(\texttt{aida-ugent/JobHop})\footnote{The dataset is available at \url{https://huggingface.co/datasets/aida-ugent/JobHop}.}. It uses an $80/10/10$ train/validation/test split over trajectories.

\paragraph{Benchmark selection.}
Of the six datasets described above, three are used as the primary benchmarks throughout this paper:
Decorte, Karrierewege, and JobHop. Together they span a wide range of scales (from $2{,}164$ to $568{,}888$ career
histories), languages (English, Dutch, and German), and annotation richness, and they represent the main publicly available
ESCO-linked career trajectory resources. The three derivative variants (Decorte-ESCO, Karrierewege-Occ,
and Karrierewege-CP) serve as supplementary robustness checks; their results are reported in
Appendix~\ref{app:additional_results}. The fourth dataset used in the main evaluation, JobHop~v2,
is an original contribution of this paper; its construction, characteristics, and improvements over the original JobHop
are described in the following subsection.

\subsection{JobHop~v2}\label{sec:JobHop_v2_dataset}

As part of this paper we introduce JobHop~v2, an improved version of the JobHop dataset~\cite{jobhop-v1}, 
derived from the same VDAB resume source
(the Flemish Public Employment Service) as the original release. After plain-text conversion and
pseudonymization, approximately $400{,}000$ multilingual (Dutch, French, and English) resumes entered the
extraction pipeline, of which $355{,}315$ remain as usable career trajectories after extraction and cleaning. All processing was performed on internal infrastructure to
satisfy the data-use agreement, and the released dataset preserves the privacy protections of the original release
(location removal and quarter-level date coarsening). Full data-source and ethics details appear in
Appendix~\ref{app:jobhop_v2_overview}.

The v2 pipeline addresses two practical limitations of the v1 release: degraded extraction quality on noisy multilingual
inputs, and failure to produce consistently machine-parseable output. Three pipeline changes are responsible for the
improvements:
\paragraph{Reasoning-controlled LLM extraction}
with \texttt{openai/gpt-oss-120b}~\cite{openai2025gptoss} at \textsc{high} reasoning effort,
paired with a retry mechanism (extended generation budget, increased repetition penalty) that recovers samples failing
initial JSON validation, yielding a $100\%$ parse rate over the ${\sim}400{,}000$ input resumes
(Appendix~\ref{app:inference});
\paragraph{Richer extraction schema}
expanding work-experience fields (standardized title, work schedule, contract type, technical skills)
and replacing v1's binary tertiary-degree flag with a five-level education taxonomy (None / Secondary / Bachelor / Master
/ PhD) (Appendix~\ref{app:extraction_prompt});
\paragraph{Multi-step normalization and cleaning}
including a two-step ESCO occupation-code assignment with asymmetric confidence thresholds,
multilingual temporal and education normalization, and consecutive-job merging that consolidates resume-formatting-induced
fragmentation (Appendices~\ref{app:normalization} and~\ref{app:cleaning}).

The released dataset contains $355{,}315$ unique career trajectories with quarter-level temporal annotations,
five-level education attainment, and ESCO-mapped occupational codes, split $80/10/10$ into training, validation,
and test sets; complete dataset statistics appear in Appendix~\ref{app:dataset_summary}, and the protocol and results of
the extraction evaluation are given in Appendices~\ref{app:evaluation_protocol} and~\ref{app:extraction_eval}, respectively.
The five-level education taxonomy and quarter-level intervals make JobHop~v2 the primary full-mode evaluation dataset for
STEP in this work, exercising both FiLM-based degree conditioning and the time-decay GRU.

\section{Evaluation}\label{sec:evaluation}

This section evaluates STEP and the embedding space ROUTE provides.
Section~\ref{sec:exp_setup} outlines the experimental details (datasets, metrics, implementation details)
shared across all experiments; Sections~\ref{sec:repr_retrieval} and~\ref{sec:prediction} report the embedding-quality and
main prediction results respectively; Section~\ref{sec:ablation_study} presents an ablation study that isolates the
contributions of STEP's individual architectural components and of the embedding stack, and
Section~\ref{sec:beyond_single} reports analyses beyond single-step accuracy; and Section~\ref{sec:pred_discussion}
synthesises the findings.

\subsection{Experimental Setup}\label{sec:exp_setup}

This subsection describes the common experimental setup that the rest of Section~\ref{sec:evaluation} instantiates:
which datasets each experiment uses, how metrics are computed, and which implementation choices are shared across models.

\textbf{Datasets.}
We evaluate on the four primary datasets introduced in Section~\ref{sec:dataset}: Decorte,
Karrierewege, JobHop, and JobHop~v2. For all experiments we adopt the official released split for each dataset rather
than resplitting, so our results are directly comparable to prior work. JobHop and JobHop~v2 provide temporal intervals and degree levels,
supporting \emph{full-mode} evaluation; Decorte and Karrierewege lack these annotations, so we use \emph{reduced-mode}
evaluation (uniform time intervals, degree set to ``None''), which serves as a controlled test of whether STEP's
architectural innovations contribute independently of the richer input features. Results on three additional derivative
datasets (Decorte-ESCO, Karrierewege-Occ, Karrierewege-CP) appear in Appendix~\ref{app:additional_results}.

\textbf{Target representation.}
Decorte is the only dataset that provides real, free-text job titles as retrieval targets;
for this dataset we use the raw job-title strings as positive targets during both training and evaluation.
The remaining three datasets (Karrierewege, JobHop, JobHop~v2) provide only ESCO occupation codes.
For these datasets, each code is resolved to its corresponding ESCO occupation description $o_k$ from the ESCO taxonomy~\cite{le2014esco}
(v1.1.2, $K = 3{,}007$ leaf-node occupations).
Retrieval at inference is therefore performed over the full ESCO occupation set $\mathcal{O} = \{o_1, \ldots, o_K\}$
using FAISS~\cite{johnson2019billion} with an inner-product index over $L_2$-normalized embeddings.

\textbf{Metrics.}
Embedding-quality experiments report Mean Reciprocal Rank (MRR), Recall@5 (R@5), and Recall@10 (R@10)
computed via cosine-similarity retrieval of the top-$k$ nearest target descriptions for each test trajectory prefix.
Prediction experiments report the same metrics over the predicted next-job distribution.
For analyses beyond single-step accuracy we additionally report Diversity@10 and a Novelty score,
defined in Section~\ref{sec:beyond_single}.

\textbf{Implementation details.}
All prediction models use 768-dimensional input embeddings. STEP uses hidden dimension 64,
degree embedding dimension 16, and dropout rate 0.1. The BiLSTM baseline uses hidden dimension 16 (32 bidirectional),
and the MLP baseline uses hidden dimension 512. All stochastic models (MLP, BiLSTM, STEP) use a training batch size of 64
and are trained with five random seeds; we report mean~$\pm$~standard deviation. On Decorte,
models are trained for up to 30 epochs with early stopping patience of 5; on all other datasets,
up to 10 epochs with patience 2. Deterministic methods (Linear, Markov, direct retrieval)
are reported as single values. Inter-job time intervals are clamped to $[0.1, 20.0]$ years to limit the influence of
extreme outliers; the upper bound of 20 years was chosen after experimenting with ten different thresholds.

\subsection{ROUTE: Embedding Quality}\label{sec:repr_setup}\label{sec:repr_retrieval}

This subsection evaluates ROUTE, the embedding space STEP relies on for retrieval and prediction.
ROUTE is the two-stage procedure (TSDAE domain adaptation followed by GISTEmbedLoss-guided contrastive fine-tuning)
defined in Section~\ref{sec:representation}; here we measure its contribution against single-stage and partial-pipeline
variants. Results on three additional derivative datasets (Decorte-ESCO, Karrierewege-Occ,
Karrierewege-CP) appear in Appendix~\ref{app:additional_results}.

We compare four embedding configurations, structured as an ablation ladder with one design choice varying per step:
(i)~\textbf{Base}\label{sec:repr_baseline}: the baseline embedding method established by prior work,
\texttt{all-mpnet-base-v2}~\cite{song2020mpnet} fine-tuned
with MultipleNegativesRankingLoss (MNRL)~\cite{henderson2017efficient}, as introduced by Decorte et
al.~\cite{decorte2023career} on their Decorte benchmark and adopted by Senger et al.~\cite{senger2025methods} in the
Karrierewege reproduction study, against which we compare our proposed embedding configurations;
(ii)~\textbf{GISTBase} (\emph{loss ablation}): same \texttt{all-mpnet-base-v2} backbone, but trained with
GISTEmbedLoss~\cite{solatorio2024gistembed} instead of MNRL, isolating the contribution of the loss function while keeping
the backbone fixed; (iii)~\textbf{NDA} (\emph{No Domain Adaptation}):
\texttt{multilingual-e5-base}~\cite{wang2024multilingual}
fine-tuned with GISTEmbedLoss only, without the TSDAE pre-training stage, isolating the effect of the multilingual
backbone while keeping the loss fixed; (iv)~\textbf{ROUTE}: \texttt{multilingual-e5-base} with TSDAE domain adaptation
followed by GISTEmbedLoss fine-tuning (Section~\ref{sec:representation}), the full proposed system.

Each transition in Table~\ref{tab:rl_direct_retrieval} isolates exactly one design choice:
Base~$\to$~GISTBase swaps the loss (MNRL $\to$ GISTEmbedLoss) while holding the backbone fixed;
GISTBase~$\to$~NDA swaps the backbone (\texttt{all-mpnet-base-v2} $\to$
\texttt{multilingual-e5-base}) while holding the loss
fixed; and NDA~$\to$~ROUTE adds TSDAE domain adaptation while holding both backbone and loss fixed.

\textbf{GISTEmbedLoss alone (Base $\to$ GISTBase) drives the largest and most consistent gains.}
Replacing MNRL with GISTEmbedLoss, while keeping the \texttt{all-mpnet-base-v2} backbone, improves performance uniformly
across all four datasets. R@10 gains are $+0.016$ (Decorte), $+0.018$ (Karrierewege), $+0.034$ (JobHop),
and $+0.018$ (JobHop~v2). Notably, GISTBase achieves the highest MRR on two of four datasets (Decorte and JobHop),
surpassing both NDA and the full ROUTE model there, indicating that guided-negative training with GISTEmbedLoss is a
primary driver of top-rank precision.

\textbf{Switching the backbone (GISTBase $\to$ NDA) yields mixed effects.}
The multilingual backbone consistently helps on Karrierewege across all metrics, but on the remaining three datasets it
reduces MRR ($-0.012$, $-0.012$, $-0.020$ on Decorte, JobHop, and JobHop~v2, respectively).
On JobHop and JobHop~v2 it nonetheless offers modest R@5 and R@10 gains, while on Decorte it also lowers R@5 and R@10.
This pattern suggests that \texttt{multilingual-e5-base} adds representational
breadth on the JobHop benchmarks (improving recall at higher $k$), whereas \texttt{all-mpnet-base-v2} with GISTEmbedLoss
already captures sufficient top-rank precision on these datasets.

\textbf{TSDAE domain adaptation (NDA $\to$ ROUTE) provides additional recall gains.}
The full ROUTE model further improves R@10 on Decorte ($+0.024$), JobHop ($+0.001$), and JobHop~v2 ($+0.019$),
achieving the highest R@10 on three of four datasets. On Karrierewege, TSDAE slightly reduces performance,
suggesting that domain adaptation is most beneficial when the pretrained embedding space is farther from the target
domain.

\begin{table*}[t]
\centering
\caption{Direct cosine-similarity retrieval performance of embedding variants on the four primary datasets.}
\label{tab:rl_direct_retrieval}
\begin{threeparttable}
\resizebox{\textwidth}{!}{%
\begin{tabular}{l ccc ccc ccc ccc}
\toprule
& \multicolumn{3}{c}{\textbf{Decorte}} & \multicolumn{3}{c}{\textbf{Karrierewege}} & \multicolumn{3}{c}{\textbf{JobHop}} & \multicolumn{3}{c}{\textbf{JobHop~v2}} \\
\cmidrule(lr){2-4} \cmidrule(lr){5-7} \cmidrule(lr){8-10} \cmidrule(lr){11-13}
\textbf{Model} & MRR & R@5 & R@10 & MRR & R@5 & R@10 & MRR & R@5 & R@10 & MRR & R@5 & R@10 \\
\midrule
Base     & 22.3 & 34.9 & 43.4 & 36.3 & 39.7 & 45.1 & 11.5 & 16.4 & 18.7 & 8.3 & 15.8 & 18.2 \\
GISTBase & \textbf{24.2} & \textbf{37.1} & 45.0 & 37.3 & 42.8 & 46.9 & \textbf{11.6} & 18.8 & 22.1 & 10.9 & 16.7 & 20.0 \\
NDA      & 23.0 & 35.5 & 43.7 & \textbf{37.9} & \textbf{43.7} & \textbf{47.6} & 10.4 & \textbf{19.0} & 22.2 & 8.9 & 17.7 & 20.8 \\
ROUTE    & 23.3 & 35.0 & \textbf{46.1} & 37.6 & 42.6 & 47.2 & 10.8 & \textbf{19.0} & \textbf{22.3} & \textbf{11.0} & \textbf{18.7} & \textbf{22.7} \\
\bottomrule
\end{tabular}%
}
\begin{tablenotes}[flushleft]
\scriptsize
\item Base: \texttt{all-mpnet-base-v2} + MNRL. Single run; direct retrieval is deterministic.
\item GISTBase: \texttt{all-mpnet-base-v2} + GISTEmbedLoss (backbone ablation; same backbone as Base, improved loss).
\item NDA: \texttt{multilingual-e5-base} + GISTEmbedLoss (no TSDAE domain adaptation).
\item ROUTE: \texttt{multilingual-e5-base} + TSDAE + GISTEmbedLoss. Bold = best per dataset and metric. All values are percentages (\%).
\end{tablenotes}
\end{threeparttable}
\end{table*}

Overall, ROUTE improves R@10 over Base on all four datasets, with the largest gains on JobHop~v2 ($+0.045$)
and JobHop ($+0.036$).
MRR shows more heterogeneous behavior: GISTEmbedLoss (captured by the GISTBase model) consistently improves top-rank
precision over Base, but adding the multilingual backbone does not preserve that advantage.
This asymmetry reflects complementary roles: GISTEmbedLoss sharpens top-rank precision, while the backbone switch and
TSDAE pretraining broaden recall. For the downstream pipeline, in which the embedding model serves as a candidate
retrieval stage feeding learned prediction models (Section~\ref{sec:prediction}), this trade-off is favourable.

\subsection{STEP: Career Path Recommendation Results}\label{sec:prediction}\label{sec:pred_results}

This subsection reports the career path recommendation results: STEP compared against five baseline prediction models on
the four primary benchmarks introduced in Section~\ref{sec:dataset}.
Section~\ref{sec:pred_baselines} describes the baselines, Section~\ref{sec:pred_main} reports the headline comparison,
Section~\ref{sec:pred_e2e} contrasts the full ROUTE+STEP pipeline with $\text{ST}_\text{Base}$+MLP and $\text{ST}_\text{Base}$+BiLSTM configurations,
and Section~\ref{sec:prior_results} situates the numbers against prior reported results in the literature.

\subsubsection{Baseline Models}\label{sec:pred_baselines}

We compare STEP against five baseline methods spanning statistical, retrieval-based, and neural approaches; the
retrieval-based Direct Retrieval baseline is evaluated in two variants ($\text{ST}_\text{Base}$ and
$\text{ST}_\text{ROUTE}$), giving six rows in Table~\ref{tab:cpr_main_v3}.
All neural baselines consume the same sentence transformer embeddings, so performance differences reflect prediction
architecture rather than input representation.

\textbf{Markov.}
A non-parametric baseline that constructs a first-order transition matrix $\mathbf{M} \in \mathbb{R}^{K \times K}$ from
training data, where $M_{ij}$ counts observed transitions from occupation $i$ to $j$. At test time,
the most recent occupation indexes into $\mathbf{M}$, and the $k$ highest-count successors are returned.
This baseline captures only first-order Markov statistics but quantifies the marginal benefit of more expressive models.

\textbf{Direct Retrieval.}
Two retrieval-only baselines apply no learned transformation: the career history embedding is used directly for
nearest-neighbor lookup. $\text{ST}_\text{Base}$ uses the Base embedding model used in Decorte et
al.~\cite{decorte2023career}, and $\text{ST}_\text{ROUTE}$ uses ROUTE two-stage domain-adapted embeddings without any
learned prediction component. The gap between these two variants isolates the representation learning contribution from
Section~\ref{sec:representation} before any learned prediction is applied.

\textbf{Linear Transformation.}
Following Decorte et al.~\cite{decorte2023career}, a least-squares transformation matrix
$\mathbf{T} \in \mathbb{R}^{d \times d}$ maps career history embeddings to the ESCO space:
$\hat{\mathbf{y}} = \mathbf{x}\mathbf{T}$, where
$\mathbf{T} = \arg\min_\mathbf{T} \|\mathbf{B} - \mathbf{A}\mathbf{T}\|_F^2$. This deterministic baseline has no
sequential component.

\textbf{MLP.}
Following Senger et al.~\cite{senger2025methods}, a two-layer feed-forward network with a residual connection:
\begin{equation}
    \hat{\mathbf{y}} = \text{normalize}\bigl(\mathbf{x} + \mathbf{W}_2\,\text{ReLU}(\mathbf{W}_1 \mathbf{x})\bigr),
\end{equation}
with $\mathbf{W}_1 \in \mathbb{R}^{512 \times d}$, $\mathbf{W}_2 \in \mathbb{R}^{d \times 512}$.
Our implementation departs from the original in its training objective: Senger et al.\ use Cosine Embedding Loss (CEL),
a pairwise loss over individual positive pairs, whereas we use cross-entropy over temperature-scaled cosine similarities
against all $K$ ESCO embeddings simultaneously, supplying a contrastive signal across the full label space.
Under identical embeddings, the cross-entropy-trained MLP consistently improves R@5 and R@10 over the CEL-trained variant
(e.g., $+6.7$~pp R@10 on Karrierewege-CP, $+4.1$~pp on Karrierewege-Occ, $+2.0$~pp on Karrierewege),
confirming this objective change as a net improvement.

\textbf{Bidirectional LSTM (BiLSTM).}
Following Senger et al.~\cite{senger2025methods}, a single-layer bidirectional LSTM~\cite{hochreiter1997long} processes
the sequence $(f_\theta(j_1), \ldots, f_\theta(j_T))$. Final hidden states from both directions are concatenated and
projected over the ESCO label set. This model captures sequential dependencies but treats all inter-job intervals as
uniform and does not use degree information.

\subsubsection{Main Comparison}\label{sec:pred_main}

Table~\ref{tab:cpr_main_v3} reports career path recommendation results using ROUTE embeddings.
STEP achieves the highest scores across all four datasets and all metrics.

\begin{table*}[t]
\centering
\caption{Career path recommendation performance with ROUTE embeddings across four benchmark datasets.}
\label{tab:cpr_main_v3}
\begin{threeparttable}
\resizebox{\textwidth}{!}{%
\begin{tabular}{l ccc ccc ccc ccc}
\toprule
& \multicolumn{3}{c}{\textbf{Decorte}} & \multicolumn{3}{c}{\textbf{Karrierewege}} & \multicolumn{3}{c}{\textbf{JobHop}} & \multicolumn{3}{c}{\textbf{JobHop~v2}} \\
\cmidrule(lr){2-4} \cmidrule(lr){5-7} \cmidrule(lr){8-10} \cmidrule(lr){11-13}
\textbf{Model} & MRR & R@5 & R@10 & MRR & R@5 & R@10 & MRR & R@5 & R@10 & MRR & R@5 & R@10 \\
\midrule
Markov                    & 7.1 & 11.0 & 16.7 & 31.8 & 44.6 & 55.9 & 20.3 & 28.3 & 35.7 & 19.6 & 26.8 & 34.2 \\
$\text{ST}_\text{Base}\dagger$   & 22.3 & 34.9 & 43.4 & 36.3 & 39.7 & 45.1 & 11.5 & 16.3 & 18.6 & 8.3  & 15.8 & 18.2 \\
$\text{ST}_\text{ROUTE}$  & 23.3 & 35.0 & 46.1 & 37.6 & 42.5 & 47.2 & 10.8 & 19.0 & 22.3 & 11.0 & 18.7 & 22.7 \\
Linear$^\dagger$          & 20.9 & 29.8 & 37.8 & 39.4 & 47.2 & 53.8 & 16.7 & 22.2 & 27.9 & 16.2 & 22.4 & 28.4 \\
MLP$^\ddagger$            & $22.6_{\pm 0.78}$ & $31.7_{\pm 0.66}$ & $40.5_{\pm 1.29}$ & $44.9_{\pm 0.03}$ & $58.7_{\pm 0.07}$ & $68.4_{\pm 0.07}$ & $20.0_{\pm 0.05}$ & $27.1_{\pm 0.13}$ & $34.9_{\pm 0.11}$ & $21.1_{\pm 0.05}$ & $29.9_{\pm 0.08}$ & $38.5_{\pm 0.07}$ \\
BiLSTM$^\ddagger$         & $22.2_{\pm 0.38}$ & $32.3_{\pm 0.81}$ & $40.8_{\pm 0.52}$ & $45.5_{\pm 0.04}$ & $59.3_{\pm 0.05}$ & $68.9_{\pm 0.03}$ & $20.9_{\pm 0.11}$ & $29.9_{\pm 0.16}$ & $38.9_{\pm 0.16}$ & $20.7_{\pm 0.19}$ & $29.6_{\pm 0.27}$ & $38.2_{\pm 0.24}$ \\
\midrule
\textbf{STEP}             & $\bm{26.4_{\pm 0.15}}$ & $\bm{37.6_{\pm 0.32}}$ & $\bm{46.6_{\pm 0.61}}$ & $\bm{46.5_{\pm 0.02}}$ & $\bm{60.7_{\pm 0.02}}$ & $\bm{70.3_{\pm 0.01}}$ & $\bm{22.2_{\pm 0.09}}$ & $\bm{31.3_{\pm 0.07}}$ & $\bm{40.4_{\pm 0.10}}$ & $\bm{22.7_{\pm 0.03}}$ & $\bm{32.3_{\pm 0.04}}$ & $\bm{41.1_{\pm 0.03}}$ \\
\bottomrule
\end{tabular}%
}
\begin{tablenotes}[flushleft]
\scriptsize
\item Decorte and Karrierewege lack temporal/degree annotations; STEP runs in \emph{reduced mode} (uniform time, no degree) on these datasets.
\item Stochastic models (MLP, BiLSTM, STEP) report mean $\pm$ std over 5 runs. Best results per dataset and metric in \textbf{bold}.
\item $\dagger$~Decorte et al.~\cite{decorte2023career}; $\ddagger$~Senger et al.~\cite{senger2025methods}.
\end{tablenotes}
\end{threeparttable}
\end{table*}

Several patterns emerge from these results.

On Decorte, STEP achieves $26.4\%$ MRR, $37.6\%$ R@5, and $46.6\%$ R@10, outperforming both the MLP ($22.6\%$ MRR,
$40.5\%$ R@10) and the $\text{ST}_\text{ROUTE}$ retrieval baseline ($23.3\%$ MRR, $46.1\%$ R@10).
On Karrierewege (reduced mode), STEP leads with R@10 of $70.3\%$, a gain of $1.4$~pp over the BiLSTM ($68.9\%$)
and $1.9$~pp over the MLP ($68.4\%$). On JobHop~v2 (full mode), STEP leads with R@10 of $41.1\%$,
ahead of the BiLSTM ($38.2\%$) and the MLP ($38.5\%$); temporal and degree conditioning contribute additional
discriminative signal on this richly annotated dataset. On JobHop (full mode), STEP again leads with R@10 of $40.4\%$,
ahead of the BiLSTM ($38.9\%$) and the MLP ($34.9\%$).

The $\text{ST}_\text{ROUTE}$ retrieval baseline achieves strong performance on Decorte ($46.1\%$ R@10)
but is surpassed by STEP ($46.6\%$); on Karrierewege and JobHop~v2, learned sequential models lead by large margins,
confirming that sequential modeling of career history adds substantial value beyond static retrieval at scale.

\subsubsection{End-to-End Pipeline Comparison}\label{sec:pred_e2e}

Table~\ref{tab:cpr_e2e} compares the full proposed pipeline (ROUTE embeddings paired with the STEP predictor)
against two off-the-shelf baseline pipelines that pair standard representation learning ($\text{ST}_\text{Base}$:
\texttt{all-mpnet-base-v2} + MNRL) with the MLP and BiLSTM predictors.
We report mean~$\pm$~standard deviation over five independent runs. The contribution of the ROUTE embeddings themselves,
isolated from the STEP predictor, is examined separately in the embedding-method ablation of
Section~\ref{sec:repr_tsdae}.

\begin{table*}[t]
\centering
\caption{End-to-end pipeline comparison: career path recommendation performance of the full ROUTE+STEP pipeline versus two off-the-shelf baseline pipelines ($\text{ST}_\text{Base}$+MLP, $\text{ST}_\text{Base}$+BiLSTM) across all four primary datasets.}
\label{tab:cpr_e2e}
\resizebox{\textwidth}{!}{%
\begin{tabular}{l ccc ccc ccc ccc}
\toprule
& \multicolumn{3}{c}{\textbf{Decorte}} & \multicolumn{3}{c}{\textbf{Karrierewege}} & \multicolumn{3}{c}{\textbf{JobHop}} & \multicolumn{3}{c}{\textbf{JobHop~v2}} \\
\cmidrule(lr){2-4} \cmidrule(lr){5-7} \cmidrule(lr){8-10} \cmidrule(lr){11-13}
\textbf{Pipeline} & MRR & R@5 & R@10 & MRR & R@5 & R@10 & MRR & R@5 & R@10 & MRR & R@5 & R@10 \\
\midrule
$\text{ST}_\text{Base}$ + MLP    & $24.7_{\pm 1.43}$ & $34.7_{\pm 2.19}$ & $43.2_{\pm 2.44}$ & $44.2_{\pm 0.04}$ & $57.1_{\pm 0.04}$ & $67.1_{\pm 0.06}$ & $19.7_{\pm 0.04}$ & $26.6_{\pm 0.04}$ & $34.2_{\pm 0.09}$ & $20.1_{\pm 0.04}$ & $28.1_{\pm 0.09}$ & $36.2_{\pm 0.06}$ \\
$\text{ST}_\text{Base}$ + BiLSTM & $22.9_{\pm 0.35}$ & $32.7_{\pm 1.10}$ & $41.8_{\pm 0.78}$ & $45.6_{\pm 0.03}$ & $59.5_{\pm 0.05}$ & $69.1_{\pm 0.06}$ & $21.3_{\pm 0.10}$ & $30.2_{\pm 0.13}$ & $39.3_{\pm 0.11}$ & $21.1_{\pm 0.19}$ & $30.1_{\pm 0.25}$ & $38.7_{\pm 0.27}$ \\
\textbf{ROUTE + STEP (ours)} & $\bm{26.4_{\pm 0.15}}$ & $\bm{37.6_{\pm 0.32}}$ & $\bm{46.6_{\pm 0.61}}$ & $\bm{46.5_{\pm 0.02}}$ & $\bm{60.7_{\pm 0.02}}$ & $\bm{70.3_{\pm 0.01}}$ & $\bm{22.2_{\pm 0.09}}$ & $\bm{31.3_{\pm 0.07}}$ & $\bm{40.4_{\pm 0.10}}$ & $\bm{22.7_{\pm 0.03}}$ & $\bm{32.3_{\pm 0.04}}$ & $\bm{41.1_{\pm 0.03}}$ \\
\bottomrule
\end{tabular}%
}
\begin{tablenotes}[flushleft]
\scriptsize
\item $\text{ST}_\text{Base}$: \texttt{all-mpnet-base-v2} + MNRL; ROUTE: \texttt{multilingual-e5-base} + TSDAE + GISTEmbedLoss.
\item Stochastic models report mean $\pm$ std over 5 runs. Best result per metric per dataset is shown in bold.
\end{tablenotes}
\end{table*}

On Decorte, ROUTE+STEP achieves the highest scores across all metrics (MRR $26.4\%$, R@5 $37.6\%$,
R@10 $46.6\%$), outperforming both off-the-shelf baseline pipelines, $\text{ST}_\text{Base}$+MLP (MRR $24.7\%$,
R@10 $43.2\%$) and $\text{ST}_\text{Base}$+BiLSTM (MRR $22.9\%$, R@10 $41.8\%$).

On the three larger benchmarks, ROUTE+STEP again leads both baselines, reaching R@10 of $70.3\%$ on Karrierewege,
$40.4\%$ on JobHop, and $41.1\%$ on JobHop~v2, ahead of $\text{ST}_\text{Base}$+MLP ($67.1\%$, $34.2\%$, $36.2\%$) and
$\text{ST}_\text{Base}$+BiLSTM ($69.1\%$, $39.3\%$, $38.7\%$). The margin over the stronger BiLSTM baseline is largest on
JobHop~v2 ($+2.4$~pp R@10), the most richly annotated dataset, where STEP's temporal and degree conditioning add
discriminative signal.

Taken together, the full ROUTE+STEP pipeline outperforms both off-the-shelf baseline pipelines on all four datasets.
We decompose this end-to-end gain into its predictor and embedding contributions in the ablation study
(Section~\ref{sec:repr_tsdae}): holding the predictor fixed at STEP, the ROUTE embeddings account for a smaller,
dataset-dependent share of the improvement, with the STEP predictor itself as the primary driver.

\subsubsection{Comparison with Prior Reported Results}\label{sec:prior_results}

On the Decorte benchmark, the best result previously reported by Decorte et al.~\cite{decorte2023career} is $43.01\%$
R@10, achieved by a hybrid model that combines their text-based CareerBERT-ALL with a skill-based component leveraging the
ESCO skills ontology. Our text-only $\text{ST}_\text{ROUTE}$ retrieval baseline reaches $46.1\%$ R@10 on the same dataset,
surpassing this result while remaining entirely within the text domain (Table~\ref{tab:cpr_main_v3}).
STEP with ROUTE embeddings further advances this to $46.6\%$ R@10, the new best result on Decorte,
achieved with a sequential model rather than direct retrieval.

On the Karrierewege benchmark, the best previously reported result is the BiLSTM of Senger et
al.~\cite{senger2025methods}. Our BiLSTM re-implementation, using ROUTE embeddings and the cross-entropy training
objective, reaches $68.9\%$ R@10; STEP further advances this to $70.3\%$, a $1.4$~pp R@10 gain over our re-implemented
BiLSTM baseline.

Senger et al.~\cite{senger2025methods} also evaluated an LLM-based approach (fine-tuned LLaMA)
and found that it generally underperforms MLP and LSTM baselines in the standardized ESCO-based prediction setting.

Models operating on entirely different datasets and evaluation protocols (CAREER~\cite{vafa2022career} and
LABOR-LLM~\cite{du2024labor}) are not directly comparable but address complementary aspects of the career prediction
problem: scale of pretraining data and natural language generation of career advice, respectively.

\subsection{Ablation Study}\label{sec:ablation_study}

This subsection isolates the contribution of individual components, at two levels: Section~\ref{sec:pred_ablation} ablates
the STEP architecture itself on JobHop~v2, and Section~\ref{sec:repr_tsdae} ablates the embedding stack, comparing
$\text{ST}_\text{Base}$, NDA, TSDAE-only, and the full ROUTE pipeline with the STEP predictor held fixed.

\subsubsection{STEP Components}\label{sec:pred_ablation}

\begin{table}[t]
\centering
\caption{Ablation study on JobHop~v2. Each row removes one component from the full model. All values are percentages; $\Delta$R@10 is the signed difference from the full model.}
\label{tab:ablation}
\begin{threeparttable}
\begin{tabular}{l cccc r}
\toprule
\textbf{Configuration} & \textbf{MRR} & \textbf{R@1} & \textbf{R@5} & \textbf{R@10} & $\boldsymbol{\Delta}$\textbf{R@10} \\
\midrule
Full model                   & 22.90 & 15.64 & 32.43 & 41.24 & n/a \\
\midrule
$-$ Temperature scaling      & 8.18  & 0.09  & 17.49 & 24.69 & $-16.55$ \\
$-$ FiLM conditioning        & 22.65 & 15.47 & 32.11 & 40.86 & $-0.38$ \\
$-$ Degree information       & 22.78 & 15.60 & 32.21 & 40.92 & $-0.32$ \\
$-$ Attention pooling        & 22.78 & 15.58 & 32.27 & 41.01 & $-0.23$ \\
$-$ Time-decay               & 22.77 & 15.51 & 32.33 & 41.08 & $-0.16$ \\
$-$ Layer normalization      & 22.87 & 15.63 & 32.39 & 41.17 & $-0.07$ \\
$-$ $\ell_2$ normalization   & 22.87 & 15.61 & 32.41 & 41.20 & $-0.04$ \\
\midrule
Minimal (GRU only)           & 22.23 & 15.12 & 31.52 & 40.40 & $-0.84$ \\
\bottomrule
\end{tabular}
\begin{tablenotes}[flushleft]
\scriptsize
\item Results from a single run with up to 10 training epochs and early stopping (patience~2), equal budget across all configurations.
\end{tablenotes}
\end{threeparttable}
\end{table}

Table~\ref{tab:ablation} reports results; the ablation reveals a clear hierarchy of component importance.
Removing learnable temperature scaling causes catastrophic degradation: R@10 drops from $41.24\%$ to $24.69\%$ and R@1
collapses to near zero ($0.09\%$). Without temperature control, raw cosine similarities bounded to $[-1, 1]$ produce
insufficiently peaked softmax distributions over the $3{,}007$-class ESCO label space, preventing concentration of
probability mass on the correct occupation and stalling effective learning.

Removing FiLM conditioning reduces R@10 by $0.38$~pp and MRR by $0.25$~pp, confirming that multiplicative modulation
captures degree--career interactions more effectively than additive concatenation. The degree-zeroing experiment further
isolates the source of this benefit: suppressing the education-level signal entirely (while keeping FiLM active)
reduces R@10 by $0.32$~pp, nearly as large as the $0.38$~pp loss from replacing FiLM itself.

Replacing attention pooling with last-hidden-state pooling reduces R@10 by $0.23$~pp, indicating that learned step
weighting of career history provides a meaningful, if modest, benefit. The time-decay GRU contributes $0.16$~pp R@10;
layer normalization and $\ell_2$ normalization each add $0.07$ and $0.04$~pp respectively.

The minimal model (GRU only) degrades by $0.84$~pp in R@10, more than twice the $0.38$~pp loss from removing any single
component, which indicates synergistic interactions that no single ablation captures.

\subsubsection{Effect of the Embedding Method}\label{sec:repr_tsdae}

We ablate the embedding stack while holding the STEP predictor fixed, comparing four configurations that progressively
build up the ROUTE representation: the off-the-shelf $\text{ST}_\text{Base}$ embeddings, the two single-stage variants
NDA (\texttt{multilingual-e5-base} + GISTEmbedLoss only, no domain adaptation) and TSDAE-only
(\texttt{multilingual-e5-base} + TSDAE only, no contrastive stage), and the full ROUTE pipeline
(\texttt{multilingual-e5-base} + TSDAE + GISTEmbedLoss). The NDA and TSDAE-only rows isolate, respectively, the
contribution of TSDAE domain adaptation (ROUTE vs.\ NDA) and of the GISTEmbedLoss contrastive stage
(ROUTE vs.\ TSDAE-only); results are reported in Table~\ref{tab:repr_embedding_ablation}.

\begin{table*}[t]
\centering
\caption{Embedding-method ablation with the STEP predictor held fixed: career path recommendation performance for four embedding configurations across the four primary datasets.}
\label{tab:repr_embedding_ablation}
\begin{threeparttable}
\resizebox{\textwidth}{!}{%
\begin{tabular}{l ccc ccc ccc ccc}
\toprule
& \multicolumn{3}{c}{\textbf{Decorte}} & \multicolumn{3}{c}{\textbf{Karrierewege}} & \multicolumn{3}{c}{\textbf{JobHop}} & \multicolumn{3}{c}{\textbf{JobHop~v2}} \\
\cmidrule(lr){2-4} \cmidrule(lr){5-7} \cmidrule(lr){8-10} \cmidrule(lr){11-13}
\textbf{Embedding} & MRR & R@5 & R@10 & MRR & R@5 & R@10 & MRR & R@5 & R@10 & MRR & R@5 & R@10 \\
\midrule
$\text{ST}_\text{Base}$ & $26.0_{\pm 0.29}$ & $37.1_{\pm 0.57}$ & $46.0_{\pm 0.40}$ & {\boldmath$46.6_{\pm 0.03}$} & {\boldmath$60.8_{\pm 0.04}$} & {\boldmath$70.4_{\pm 0.03}$} & $22.3_{\pm 0.02}$ & $31.0_{\pm 0.08}$ & $39.9_{\pm 0.04}$ & $22.8_{\pm 0.03}$ & $32.3_{\pm 0.02}$ & $41.0_{\pm 0.03}$ \\
NDA                     & $26.3_{\pm 0.21}$ & {\boldmath$37.7_{\pm 0.90}$} & {\boldmath$46.7_{\pm 0.91}$} & $46.5_{\pm 0.03}$ & $60.6_{\pm 0.03}$ & $70.3_{\pm 0.02}$ & $22.4_{\pm 0.02}$ & $31.6_{\pm 0.05}$ & $40.6_{\pm 0.05}$ & $22.6_{\pm 0.02}$ & $32.2_{\pm 0.07}$ & $40.9_{\pm 0.10}$ \\
TSDAE-only              & $23.8_{\pm 0.62}$ & $34.6_{\pm 1.00}$ & $43.5_{\pm 0.19}$ & {\boldmath$46.6_{\pm 0.02}$} & {\boldmath$60.8_{\pm 0.02}$} & {\boldmath$70.4_{\pm 0.03}$} & {\boldmath$22.7_{\pm 0.03}$} & {\boldmath$32.1_{\pm 0.05}$} & {\boldmath$41.4_{\pm 0.06}$} & {\boldmath$22.9_{\pm 0.03}$} & {\boldmath$32.4_{\pm 0.02}$} & {\boldmath$41.3_{\pm 0.05}$} \\
\textbf{ROUTE (ours)}   & {\boldmath$26.4_{\pm 0.15}$} & $37.6_{\pm 0.32}$ & $46.6_{\pm 0.61}$ & $46.5_{\pm 0.02}$ & $60.7_{\pm 0.02}$ & $70.3_{\pm 0.01}$ & $22.2_{\pm 0.09}$ & $31.3_{\pm 0.07}$ & $40.4_{\pm 0.10}$ & $22.7_{\pm 0.03}$ & $32.3_{\pm 0.04}$ & $41.1_{\pm 0.03}$ \\
\bottomrule
\end{tabular}%
}
\begin{tablenotes}[flushleft]
\scriptsize
\item All rows use the STEP predictor; only the embedding stack varies.
\item $\text{ST}_\text{Base}$: \texttt{all-mpnet-base-v2} + MNRL; NDA: \texttt{multilingual-e5-base} + GISTEmbedLoss (no domain adaptation); 
\item TSDAE-only: \texttt{multilingual-e5-base} + TSDAE (no contrastive stage); ROUTE: \texttt{multilingual-e5-base} + TSDAE + GISTEmbedLoss.
\item Stochastic models report mean $\pm$ std over 5 runs. Best value per dataset and metric in \textbf{bold}.
\end{tablenotes}
\end{threeparttable}
\end{table*}

With the predictor fixed to STEP, the four embedding configurations differ little: on the three large benchmarks all of
them fall within ${\sim}1.5$~pp on every metric, and on Karrierewege within $0.1$~pp. Outside Karrierewege the off-the-shelf
$\text{ST}_\text{Base}$ embeddings are never the best configuration, so some domain-specific representation always helps,
but only by a small margin. On Karrierewege itself the four configurations are so close that $\text{ST}_\text{Base}$ ties
with the others for the best score on every metric, so even there the domain-specific variants offer no meaningful
advantage.

Notably, what helps \emph{direct retrieval} does not all carry over to STEP. In the retrieval setting, the
GISTEmbedLoss contrastive stage is the single largest driver of embedding quality
(Section~\ref{sec:repr_retrieval}, Table~\ref{tab:rl_direct_retrieval}). Under STEP, however, that advantage all but
vanishes on the three large benchmarks: removing the stage entirely (TSDAE-only) matches or slightly exceeds the full
ROUTE pipeline on Karrierewege, JobHop, and JobHop~v2. The reason is that STEP relearns its own scoring over the
candidate set, so it barely benefits from an embedding space that was further tuned for nearest-neighbour retrieval.

The two ROUTE stages in fact prove complementary across dataset scales. On the large benchmarks the denoising TSDAE stage
carries the benefit and the contrastive stage is redundant, whereas on Decorte, the smallest benchmark and the only one
with free-text rather than ESCO retrieval targets, the reverse holds: the contrastive stage is essential and TSDAE-only
collapses from ${\sim}46.6\%$ to $43.5\%$ R@10. ROUTE combines both stages and is therefore the safest default, the
strongest \emph{retrieval} space (Section~\ref{sec:repr_retrieval}) and competitive with the best STEP configuration on
every dataset; a practitioner who knows their setting could instead keep only the stage that helps it, TSDAE-only for
large in-domain corpora and the contrastive stage for small or free-text ones.

Either way, the embedding pipeline is a secondary lever once STEP is in place: STEP, not the choice of embedding, is the
primary driver of the end-to-end gains reported in Section~\ref{sec:pred_e2e}.

\subsection{Beyond Single-Step Accuracy}\label{sec:beyond_single}

Standard accuracy metrics measure whether the correct next job appears in the top-$k$ recommendations at a single
prediction step.
Two further dimensions of model behavior are relevant for downstream use: whether the top-$k$ list spans diverse
occupational domains and represents non-trivial moves from the current role (Section~\ref{sec:disc_diversity}),
and how performance degrades when the model is rolled out autoregressively over multiple future steps
(Section~\ref{sec:pred_multistep}).

\subsubsection{Diversity and Novelty}\label{sec:disc_diversity}

We quantify diversity and novelty over the top-10 recommendations of each model via two metrics.

\textbf{Diversity@10} reports the average number of unique ESCO 2-digit groups across the top-10 recommendations per test instance, where a score of~1 indicates all recommendations fall within a single occupational sector and higher values indicate broader cross-sector coverage.

\textbf{Novelty} is the average ESCO hierarchy distance between the last observed occupation and each recommended occupation, indicating whether recommendations represent incremental or exploratory moves.
For instance, if a worker's last observed occupation is
\textit{software developer} (ISCO~2512),
a top-1 recommendation of \textit{web developer} (ISCO~2513), sharing the same 3-digit minor group,
yields a distance of~1, whereas a recommendation of
\textit{sales manager} (ISCO~1221),
belonging to a different major group entirely, yields the maximum distance of~4.

\begin{table*}[t]
\centering
\caption{Diversity@10 and Novelty for four representative models across the four primary datasets, with the ground-truth Novelty of each test set as a reference.}
\label{tab:diversity_novelty}
\begin{threeparttable}
\begin{tabular}{l cccc cccc}
\toprule
& \multicolumn{4}{c}{\textbf{Diversity@10}} & \multicolumn{4}{c}{\textbf{Novelty}} \\
\cmidrule(lr){2-5} \cmidrule(lr){6-9}
\textbf{Model} & Dec. & Kar. & JH & JH~v2 & Dec. & Kar. & JH & JH~v2 \\
\midrule
Markov                    & 7.00 & 6.17 & 6.01 & 5.97 & 3.21 & 0.35 & 0.93 & 0.81 \\
$\text{ST}_\text{ROUTE}$ & 4.52 & 5.34 & 3.08 & 3.24 & 0.76 & 0.22 & 0.04 & 0.26 \\
BiLSTM                   & 5.59 & 6.01 & 5.43 & 5.44 & 2.20 & 0.62 & 1.83 & 1.93 \\
\textbf{STEP}            & 4.78 & 5.94 & 5.35 & 5.29 & 1.67 & 0.56 & 1.51 & 1.66 \\
\midrule
Ground truth             & --   & --   & --   & --   & 2.67 & 2.21 & 2.92 & 2.95 \\
\bottomrule
\end{tabular}
\begin{tablenotes}[flushleft]
\scriptsize
\item Dec.\ = Decorte, Kar.\ = Karrierewege, JH = JobHop, JH~v2 = JobHop~v2.
\item Diversity@10 is undefined for the single ground-truth label and is shown as ``--''.
\end{tablenotes}
\end{threeparttable}
\end{table*}

Two findings emerge from Table~\ref{tab:diversity_novelty}:

\begin{itemize}
  \item \textbf{Diversity varies substantially across models and datasets, revealing a diversity--accuracy tradeoff.}
  Markov, which returns globally frequent transitions, spans $5.97$--$7.00$ unique 2-digit groups per top-10 list,
  providing a ceiling for comparison.
  $\text{ST}_\text{ROUTE}$ scores substantially lower on both JobHop datasets ($3.08$ and $3.24$ groups),
  nearly half the breadth of Markov.
  This reflects the geometry of the embedding space: career history prefixes already reside near the current occupation in
  ESCO embedding space, so cosine nearest-neighbor lookup returns semantically similar (and therefore occupationally
  proximate) descriptions clustered within a narrow sector. STEP's learned prediction head substantially escapes this
  regime, reaching $5.35$ groups on JobHop and $5.29$ on JobHop~v2, an increase of $2.27$ and $2.05$ unique groups over
  $\text{ST}_\text{ROUTE}$, respectively. BiLSTM achieves slightly higher Diversity@10 than STEP on all four datasets,
  with the largest gap on Decorte ($5.59$ vs.\ $4.78$) and the smallest on Karrierewege ($6.01$ vs.\ $5.94$);
  on the JobHop datasets the gap is $5.43$/$5.44$ vs.\ $5.35$/$5.29$.
  This narrow diversity gap between BiLSTM and STEP on most datasets coincides with STEP's substantial accuracy advantage
  (Table~\ref{tab:cpr_main_v3}), illustrating that a modest reduction in distributional breadth is a small cost for
  substantially better predictive fidelity.

  \item \textbf{Novelty reveals a systematic conservatism in all learned models.}
  The Ground-truth row of Table~\ref{tab:diversity_novelty} reports the average ESCO distance between the last observed
  occupation and the next occupation actually held by each test individual, providing a target value against which model
  Novelty should be read. True transitions are far from local on every dataset: Decorte ($2.67$),
  JobHop ($2.92$), and JobHop~v2 ($2.95$) all sit close to the ``different major group'' end of the ISCO hierarchy,
  and even Karrierewege ($2.21$), while the least exploratory of the four, still corresponds to typically cross-sector
  moves.

  Against this reference, every learned model substantially undershoots. STEP reaches at most $1.67$ on Decorte (gap of
  $1.00$ vs.\ ground truth) and falls to $1.51$/$1.66$ on the JobHop datasets (gaps of $1.41$/$1.29$).
  The gap is largest on Karrierewege, where STEP's Novelty of $0.56$ trails the ground truth by $1.65$,
  indicating that the dataset's relatively low exploratory ceiling does not on its own explain the model's near-local
  recommendations, the model is more conservative than the data demands. BiLSTM follows the same pattern with marginally
  less collapse on JobHop/JobHop~v2 ($1.83$/$1.93$, still $\sim$$1$~pp below ground truth), while $\text{ST}_\text{ROUTE}$
  collapses most severely, with near-zero Novelty on JobHop and JobHop~v2 ($0.04$--$0.26$): top-1 retrieval returns
  occupations essentially identical to the current one.
\end{itemize}

This systematic gap clarifies STEP's accuracy gains over $\text{ST}_\text{ROUTE}$ on the JobHop datasets (R@10:
$40.4\%$ vs.\ $22.3\%$ on JobHop; $41.1\%$ vs.\ $22.7\%$ on JobHop~v2): they stem not only from sequential modeling,
but also from the ability to recommend occupations at greater ESCO distance, closer to where individuals actually
transition. Markov is the only model that exceeds ground-truth Novelty on Decorte ($3.21$ vs.\ $2.67$),
which is expected, but on Karrierewege, JobHop, and JobHop~v2 it too lies well below the ground-truth target.
Taken together, the Ground-truth row reframes Novelty as a measure of how much each model leaves on the table:
a non-trivial fraction of the true transitions corresponds to genuine cross-domain moves that none of the evaluated models
currently recover.

\subsubsection{Multi-Step Autoregressive Evaluation}\label{sec:pred_multistep}

We evaluate STEP in an autoregressive multi-step setting and measure how performance degrades as the prediction horizon
$M$ grows from~1 to~5.

\paragraph{Protocol.}
For each test trajectory of length $L$ and each valid starting position $T$ (where $L - T \geq M_{\max}$),
we (i)~encode the first $T$ jobs as the input prefix, (ii)~run STEP to predict the next occupation via nearest-neighbor
retrieval in the ESCO embedding space, (iii)~append the top-1 predicted embedding to the prefix (time delta set to
$1.0$~year), and (iv)~repeat for steps $m = 1, \ldots, M_{\max}$. At each horizon $m$ we record Recall@$K$
($K \in \{1, 5, 10\}$) and MRR against the true occupation at position $T + m - 1$. The protocol is compatible with the
multi-step benchmarks of NAOMI~\cite{yamashita2022looking} and CAPER~\cite{lee2025caper}. We evaluate on the three large
datasets that have sufficient trajectory depth: Karrierewege ($N = 46{,}393$ test instances),
JobHop~v2 ($N = 36{,}866$), and JobHop ($N = 6{,}498$). Note that the $M_{\max}$ filter selects only trajectories with
$L - T \geq 5$, a longer-history subset of each test set, so multi-step numbers at $M = 1$ are not directly comparable to
the single-step numbers in Table~\ref{tab:cpr_main_v3}; the gap is most visible on Karrierewege (R@10 $80.2\%$ here vs.
$70.3\%$ in Table~\ref{tab:cpr_main_v3}) and minimal on the JobHop datasets where short and long histories perform
similarly.

\begin{figure*}[t]
  \centering
  \begin{subfigure}[t]{0.32\linewidth}
    \includegraphics[width=\linewidth]{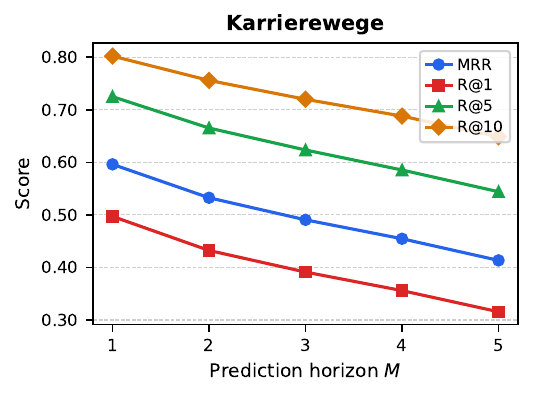}
    \caption{Karrierewege ($N=46{,}393$). STEP maintains R@10 above $64\%$ and MRR above $41\%$ at $M=5$.}
    \label{fig:multistep_karrierewege}
  \end{subfigure}
  \hfill
  \begin{subfigure}[t]{0.32\linewidth}
    \includegraphics[width=\linewidth]{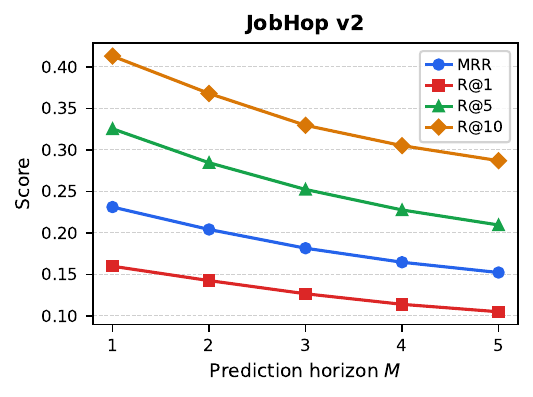}
    \caption{JobHop~v2 ($N=36{,}866$). R@10 degrades from $41.3\%$ at $M=1$ to $28.7\%$ at $M=5$; MRR from $23.1\%$ to $15.2\%$.}
    \label{fig:multistep_jobhop_v2}
  \end{subfigure}
  \hfill
  \begin{subfigure}[t]{0.32\linewidth}
    \includegraphics[width=\linewidth]{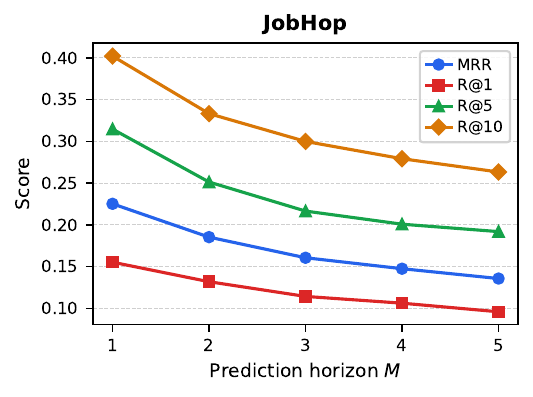}
    \caption{JobHop ($N=6{,}498$). Closely mirrors JobHop~v2, with R@10 from $40.2\%$ ($M=1$) to $26.3\%$ ($M=5$).}
    \label{fig:multistep_jobhop}
  \end{subfigure}
  \caption{Multi-step autoregressive evaluation across all three datasets. Performance degrades monotonically with prediction horizon $M$, but gradually rather than catastrophically.}
  \label{fig:multistep_all}
\end{figure*}

\paragraph{Results.}
Figures~\ref{fig:multistep_karrierewege}--\ref{fig:multistep_jobhop} show MRR, R@1, R@5, and R@10 as a function of
prediction horizon for the three datasets. Performance degrades monotonically with horizon on all datasets,
as expected from error accumulation in autoregressive rollout, yet the degradation is gradual rather than catastrophic.

On Karrierewege, STEP achieves R@10 of $80.2\%$ at $M=1$, declining to $64.9\%$ at $M=5$, an absolute drop of $15.3$~pp
over five autoregressive steps.
MRR follows a similar trajectory, falling from $59.6\%$ to $41.3\%$. The modest degradation on this dataset is consistent
with its high single-step accuracy: the autoregressive prefix is often accurate, limiting error accumulation.

On JobHop and JobHop~v2, single-step performance is lower ($40.2\%$ and $41.3\%$ R@10, respectively),
and the multi-step curves are correspondingly lower throughout. Despite this, the model retains meaningful recall at
$M=5$: R@10 of $26.3\%$ (JobHop) and $28.7\%$ (JobHop~v2), substantially above a random baseline.
The two JobHop variants show nearly identical multi-step profiles, confirming that the improved extraction quality in
JobHop~v2 does not qualitatively alter the multi-step dynamics.

Across all three datasets, R@10 degrades more slowly than MRR and R@1, suggesting that while the model loses precision at
the top rank, it retains broader recall over extended horizons.

\subsection{Discussion}\label{sec:pred_discussion}

This subsection synthesises what the empirical results in Sections~\ref{sec:repr_retrieval}--\ref{sec:beyond_single}
reveal about STEP, the role of the embedding space ROUTE provides, and the broader implications for career path
recommendation.

\paragraph{Why STEP works.}
The component ablation (Section~\ref{sec:pred_ablation}, Table~\ref{tab:ablation}) is informative not only for which
parts of STEP matter but for what their ordering implies. Learnable temperature scaling dominates by a wide margin---an
expected consequence of scoring against a $3{,}007$-class label space, where similarity logits must be sharply calibrated
before any single occupation can win the softmax. More revealing is the degree pathway: decomposing FiLM conditioning
shows that nearly all of its benefit comes from the education-level signal itself rather than from the multiplicative
modulation mechanism, so incorporating educational attainment is the substantive lever and the conditioning mechanism is a
secondary design choice. Time decay, by contrast, appears marginal in the JobHop~v2 ablation taken alone,
but that single-dataset view understates it: the larger margin STEP enjoys on the full-mode datasets over the reduced-mode
ones (Table~\ref{tab:cpr_main_v3}) indicates temporal decay contributes more when genuine inter-job intervals are
available than the in-isolation ablation can reveal.

\paragraph{The role of representation quality.}
The embedding-method ablation (Table~\ref{tab:repr_embedding_ablation}) shows that the representation refinements which
most improve direct retrieval do not all propagate to STEP: the GISTEmbedLoss contrastive stage drives the largest
retrieval-quality gains (Section~\ref{sec:repr_retrieval}), yet removing it leaves STEP unchanged or marginally better on
the three large benchmarks. STEP relearns its own scoring over the candidate set, recovering the discriminative structure
that nearest-neighbour retrieval must otherwise read directly off the embedding. The two ROUTE stages are moreover
complementary: denoising domain adaptation (TSDAE) is what helps on the large in-domain benchmarks, while the contrastive
stage is what matters on the small, free-text Decorte set. ROUTE bundles both for robustness, but this modularity is
itself a practical finding: a deployment that knows its data regime can adopt only the component that benefits it without
sacrificing most of STEP's accuracy.

\paragraph{STEP relative to prior reported results.}
On Decorte, STEP sets a new SOTA while remaining entirely within the text domain, without the skill-based components used
by the prior best hybrid result of Decorte et al.~\cite{decorte2023career}; skill components are not used by STEP and
remain a complementary direction for further gains. That this is achieved with a sequential model on a small benchmark is
itself notable: with adequate training budget (batch size $64$, $30$ epochs, patience~$5$),
sequential architectures do not inherently underperform simpler ones at this scale, a counterpoint to a recurring
assumption in the literature. On Karrierewege, STEP's gain over the re-implemented BiLSTM is conservative---the gap
against the original Senger et al.~\cite{senger2025methods} numbers would be larger, since our baseline already
incorporates improved embeddings. The finding that fine-tuned LLMs underperform MLP and LSTM baselines in this
setting~\cite{senger2025methods} also fits our framing: retrieval over a fixed ESCO label space is structurally different
from next-token generation, and a model optimised for the latter offers no clear advantage over a specialised retrieval
architecture.

\paragraph{Predictive versus exploratory recommendation.}
The diversity and novelty analysis in Section~\ref{sec:disc_diversity} reveals a meaningful distinction between two modes
of deployment. A system optimised purely for accuracy gravitates toward short-distance recommendations:
STEP's high accuracy on Karrierewege ($70.3\%$ R@10) coincides with the lowest predicted Novelty ($0.56$),
and Karrierewege is also the dataset whose ground-truth transitions are the most local of the four (Ground-truth Novelty
$2.21$). Yet even this most-local dataset has a ground-truth Novelty nearly four times STEP's predicted value,
and on JobHop and JobHop~v2 the gap between predicted ($\sim$$1.5$--$1.7$) and ground-truth Novelty ($\sim$$2.9$)
is similarly large. In other words, models that score well on accuracy-style metrics systematically underestimate how
exploratory true career moves are; a career-counseling application aimed at supporting occupational exploration would need
to explicitly promote higher-novelty recommendations, which the current STEP objective does not do.
The diversity--accuracy trade-off between the BiLSTM (slightly broader top-10 lists) and STEP (substantially higher
accuracy) further illustrates that the choice depends on the application; reconciling predictive fidelity with exploratory
breadth is a productive direction for future work.

\paragraph{Multi-step generalisation.}
The autoregressive multi-step evaluation in Section~\ref{sec:pred_multistep} shows STEP's degradation with horizon is
gradual rather than catastrophic, with R@10 declining from $80.2\%$ to $64.9\%$ on Karrierewege over five steps and
remaining substantially above random on the JobHop datasets at $M=5$. That R@10 degrades more slowly than MRR and R@1
means STEP loses precision at the top rank but retains broader recall over extended horizons---a favourable property for
career-counseling applications, where a shortlist of plausible future occupations is more useful than a committed single
prediction.

\paragraph{Data handling and privacy.}
Data use is authorized under a formal research agreement with VDAB and conducted in compliance with applicable ethical and
legal requirements. All processing was performed on local infrastructure; resume contents were never transmitted to
external services. The released dataset incorporates multi-layered privacy protections: VDAB pseudonymization,
on-premises processing, removal of location fields, and coarsening of all dates to quarter-level granularity.

\paragraph{Limitations and ethical considerations.}
Career prediction systems carry inherent risks. A model trained on historical career trajectories inevitably learns
patterns that reflect historical inequities in labour-market access, occupational segregation by gender or socioeconomic
status, and systemic biases in hiring decisions~\cite{de-artega2019bias}. Deploying such a model without fairness
evaluation risks reinforcing existing structural inequalities by steering individuals toward occupations they are
historically likely to enter rather than those they could succeed in given appropriate support~\cite{salinas2023unequal}.
ESCO standardisation provides a neutral occupational vocabulary but does not eliminate the biases encoded in the
trajectories used for training, and the current STEP objective does not include any fairness-aware term.
Scale introduces a further caveat: the causal benefits of occupational advice can attenuate, and even reverse through
displacement, when identical recommendations are issued to many competing job seekers at once~\cite{altmann2022online},
so personalised guidance should be evaluated as deployed rather than assumed to transfer from individual-level gains.
We strongly recommend that any deployment of this pipeline be accompanied by rigorous fairness evaluation across
demographic groups before it is used in career counseling or hiring support applications.

\section{Conclusion}\label{sec:conclusion}

This paper presented STEP, a sequential model for next job prediction in career paths that integrates a time-decay GRU,
FiLM-based degree conditioning, attention-based sequence pooling, and learnable temperature scaling over a domain-adapted
ESCO embedding space. STEP attains state-of-the-art R@10 on all four primary benchmarks, including $46.6\%$ on
Decorte---surpas

\begin{acks}
This research was funded by the BOF of Ghent University (BOF20/IBF/117),
the Flemish Government (AI Research Program), the FWO (G073924N),
and the EU (ERC, VIGILIA, 101142229).
Views and opinions expressed are however those of the author(s) only and do not necessarily reflect those of the EU or the ERC Executive Agency.
Neither the EU nor the granting authority can be held responsible for them.
For the purpose of Open Access the author has applied a CC BY public copyright license to any Author Accepted Manuscript version arising from this submission.
Part of the experiments were conducted on pseudonimized HR data generously provided by VDAB.
\end{acks}

\bibliographystyle{ACM-Reference-Format}
\bibliography{bibliography}


\appendix

\section*{Overview of the Appendices}

The appendices collect the supplementary technical material supporting the main paper, organized into three groups.
The first group documents the construction of JobHop~v2: its data source and ethical framework
(Appendix~\ref{app:jobhop_v2_overview}), the LLM extraction prompt (Appendix~\ref{app:extraction_prompt}),
the inference infrastructure and throughput (Appendix~\ref{app:inference}), the data cleaning pipeline
(Appendix~\ref{app:cleaning}), the normalization pipeline including ESCO occupation-code assignment
(Appendix~\ref{app:normalization}), and a summary of the resulting dataset (Appendix~\ref{app:dataset_summary}).
It also reports how extraction quality was assessed, covering the evaluation protocol
(Appendix~\ref{app:evaluation_protocol}) and the labeled datasets and results
(Appendix~\ref{app:extraction_eval}).
The second group concerns the ROUTE representation model: the exploration of candidate embedding backbones and training
configurations (Appendix~\ref{app:embedding_exploration}) and the final fine-tuning hyperparameters
(Appendix~\ref{app:embedding_hyperparams}).
The third group concerns the STEP recommendation model: its architecture and hyperparameters
(Appendix~\ref{app:model_architecture}) and additional career-path recommendation results on the derivative
benchmark variants (Appendix~\ref{app:additional_results}).

\newcommand{\hlmask}{\colorbox{orange!40}{\ttfamily\tiny<MASK>}}
\newcommand{\jkey}[1]{\textcolor{blue!60!black}{"#1"}}
\newcommand{\jval}[1]{\textcolor{teal!55!black}{"#1"}}
\newcommand{\jnote}[1]{\textcolor{gray!70}{\normalfont\scriptsize\textit{\ // #1}}}

\begin{figure*}[t!]
\centering
\begin{minipage}[t]{0.54\textwidth}
  \vspace{0pt}
  {\small\bfseries Extraction Rules (Sections 1--3)}\\[1pt]
  \begin{tcolorbox}[
    enhanced, colback=blue!2, colframe=blue!30,
    boxrule=0.5pt, arc=3pt,
    left=6pt, right=5pt, top=5pt, bottom=5pt,
    fontupper=\normalfont
  ]
  {\scriptsize\ttfamily
You are a strict Resume Parsing Engine. Your goal is to extract structured
data from the resume provided below into a single valid JSON object.

\medskip
\noindent\textbf{1.\ GENERAL CONSTRAINTS}\\
- \textbf{Input:} Resume text captured at timestamp: \{timestamp\}.\\
- \textbf{Output:} A single JSON object. No markdown, no explanations, no <MASK> tokens.\\
- \textbf{Language:} Do NOT translate descriptions or titles. Keep original language.\\
- \textbf{Missing Data:} Use ``-'' for missing/masked values. Drop parts containing only <MASK>.\\
- \textbf{Current Roles:} Replace markers like ``Present'', ``Heden'', ``Current'' with the\\
\quad provided timestamp formatted as ``Month Year''.

\medskip
\noindent\textbf{2.\ LOGIC \& PARSING RULES}\\
- \textbf{MERGE:} Consecutive entries for the same role/company with overlapping or adjacent\\
\quad dates must be combined into one entry.\\
- \textbf{SPLIT:} Single entries listing disjoint dates (e.g., ``2008, 2009 and 2015'') must be\\
\quad split into separate entries for each period.\\
- \textbf{GROUPS:} Unpack grouped sections (e.g., ``Interims'', ``Stages'') into individual experiences.\\
- \textbf{Dates:}\\
\quad Format: ``Month Year'' or ``Year''.\\
\quad Do not invent months. If only a year is listed, output only the year.\\
\quad Handle ranges: ``May July 2013'' $\to$ Start: ``May 2013'', End: ``July 2013''.\\
\quad Handle masks: ``<MASK>-2009'' $\to$ Start: ``-'', End: ``2009''.\\
\quad ``<MASK>/<MASK>/2015--<MASK>/<MASK>/2018'' $\to$ Start: ``2015'', End: ``2018''.\\
\quad \textbf{CLEANUP:} Ensure no hyphens or slashes remain from the mask.\\
\quad Always extract visible years.\\
- \textbf{Mappings:}\\
\quad Internships: ``stage'', ``trainee'', ``leerling'' $\to$ \texttt{contract\_type}: ``internship''.\\
\quad Degrees: ``Hogeschool'' $\to$ ``Bachelor''; ``Lic.'' $\to$ ``Master''; ``Doctoraat'' $\to$ ``PhD''.\\
\quad Languages: ``moedertaal/native'' $\to$ ``native''; ``vloeiend/fluent'' $\to$ ``fluent'';\\
\quad\quad ``goed'' $\to$ ``intermediate''; ``basis/notions'' $\to$ ``basic''.

\medskip
\noindent\textbf{3.\ EXTRACTION SCOPE}\\
- \textbf{Skills:} Extract ONLY technical skills/tools. Ignore soft skills.\\
- \textbf{Experience:}\\
\quad CRITICAL: Do NOT extract items from ``Education'' or ``Profile'' sections as work experience.\\
\quad Be careful with multi-column layouts.\\
\quad For the \texttt{title} field, extract the job title exactly as written.\\
\quad For the \texttt{standard\_title} field, infer a standardized title (preferably ESCO).\\
- \textbf{Certificates:} professional training/courses only.

  }
  \end{tcolorbox}
\end{minipage}
\hfill
\begin{minipage}[t]{0.45\textwidth}
  \vspace{0pt}
  {\small\bfseries Output Schema}\\[1pt]
  \begin{tcolorbox}[
    enhanced, colback=blue!2, colframe=blue!30,
    boxrule=0.5pt, arc=3pt,
    left=6pt, right=5pt, top=5pt, bottom=5pt
  ]
  {\scriptsize\ttfamily\textbf{4.\ OUTPUT SCHEMA}}\\[2pt]
  {\scriptsize\begin{alltt}
\{
  \jkey{work\_experiences}: [
    \{
      \jkey{title}: "Original Job Title",
      \jkey{standard\_title}: "Standardized Job Title",
      \jkey{company\_name}: "Organization Name (or '-')",
      \jkey{location}: "City only (no country)",
      \jkey{description}: "Original text description",
      \jkey{start\_date}: "Month Year | Year | -",
      \jkey{end\_date}: "Month Year | Year | -",
      \jkey{work\_schedule\_type}:
        "full-time | part-time | -",
      \jkey{contract\_duration}:
        "temporary | permanent | seasonal | -",
      \jkey{contract\_type}:
        "employment | freelance | internship |
         volunteer | student job | -",
      \jkey{skills\_used}: ["Skill 1", "Skill 2"]
    \}
  ],
  \jkey{educations}: [
    \{
      \jkey{degree\_level}:
        "Primary education | Secondary school |
         Bachelor | Master | PhD",
      \jkey{degree\_title}: "Title (e.g. B.Sc. IT)",
      \jkey{institute\_name}: "School/University Name",
      \jkey{start\_date}: "Month Year | Year",
      \jkey{end\_date}: "Month Year | Year"
    \}
  ],
  \jkey{languages}: [
    \{
      \jkey{language}: "Name",
      \jkey{proficiency\_level}:
        "native | fluent | intermediate | basic"
    \}
  ],
  \jkey{other\_certificates}: [
    \{
      \jkey{certificate\_name}: "Name",
      \jkey{issuing\_organization}: "Issuer Name",
      \jkey{issue\_date}: "Month Year | Year"
    \}
  ]
\}
  \end{alltt}}
  \end{tcolorbox}
\end{minipage}
\caption{%
  Full developer instruction used for resume extraction (Section~\ref{sec:JobHop_v2_dataset}).
  \textbf{Left:}~Extraction rules covering general constraints (\S1), date-handling logic
  and type-mapping conventions (\S2), and extraction scope (\S3).
  \textbf{Right:}~Target JSON output schema; field names shown in
  \textcolor{blue!60!black}{blue}.
  Curly braces denote JSON structure; \texttt{\{timestamp\}} is replaced at runtime
  with the capture date.
}
\label{fig:extraction_prompt}
\end{figure*}

\section{JobHop~v2: Data Source and Ethical Framework}\label{app:jobhop_v2_overview}

This section expands on the JobHop~v2 dataset introduced in Section~\ref{sec:JobHop_v2_dataset}. It documents the data source and the ethical and legal framework governing its use.

\subsection*{Data Source}

JobHop~v2 is derived from a corpus of ${\sim}440{,}000$ pseudonymized resumes provided by VDAB,
the Flemish Public Employment Service, under a formal research collaboration agreement. Prior to sharing,
VDAB converted all resumes to plain text and applied systematic pseudonymization: personally identifiable information and
low-frequency terms were replaced with
\texttt{<MASK>} tokens to reduce indirect re-identification risk. The collection spans multiple
languages (predominantly Dutch, with smaller portions in French and English), reflecting the multilingual labor market of
the Flemish region of Belgium.

Not all ${\sim}440{,}000$ source resumes were processable after plain-text conversion and pseudonymization:
documents whose formatting produced empty or near-empty text after VDAB's transformation pipeline were excluded prior to
extraction, yielding approximately $400{,}000$ input resumes that entered the LLM extraction stage
(Appendix~\ref{app:inference}).

\subsection*{Ethical and Legal Framework}

Data use is governed by the research agreement and complies with applicable ethical and legal requirements.
A binding constraint requires all processing to occur on internal infrastructure: resume contents are never transmitted to
external services or third-party APIs. This constraint shaped the extraction architecture (Appendix~\ref{app:inference}),
which relies exclusively on locally hosted model inference. The released dataset incorporates additional privacy
protections: location information is removed and all dates are coarsened to quarter-level granularity,
ensuring no personally identifiable information remains in any public release.

\section{LLM Extraction Prompt}\label{app:extraction_prompt}

The extraction prompt is one of the main design choices in the production pipeline (Section~\ref{sec:JobHop_v2_dataset}).
The prompt has two components. The \textbf{system message} is a single sentence establishing the model's role:
\textit{``You are an expert HR assistant.
Extract only what is explicitly present and format as valid JSON.''} The \textbf{developer instruction} contains the full
rule set and output schema, reproduced in Figure~\ref{fig:extraction_prompt}.
In Figure~\ref{fig:extraction_prompt}, internal infrastructure details and model-specific encoding directives have been abstracted;
the complete extraction logic, output schema, and all parsing rules are preserved verbatim.

The prompt is injected into the model's conversation using the Harmony encoding framework,
which provides explicit reasoning-effort control via the system message. The resume text is appended as a user message
after the developer instruction.

Figure~\ref{fig:extraction_example} illustrates the end-to-end behavior of the extraction pipeline on a synthetic resume
representative of the VDAB corpus.

\begin{figure*}[t!]
\centering
\begin{minipage}[t]{0.44\textwidth}
  \vspace{0pt}
  {\small\bfseries (a) Pseudonymized resume (input)}\\[1pt]
  \begin{tcolorbox}[
    enhanced, colback=gray!4, colframe=gray!50,
    boxrule=0.5pt, arc=3pt,
    left=6pt, right=5pt, top=5pt, bottom=5pt,
    lower separated=false,
    colbacklower=gray!8,
    fontupper=\normalfont
  ]
  {\scriptsize\begin{alltt}
\textbf{Curriculum Vitae}

\hlmask{} \hlmask{}
\hlmask{}, 9000 Gent
\hlmask{}@\hlmask{}.be   +32 \hlmask{}
Belgisch, \textdegree{}\hlmask{}/1988

\textbf{WERKERVARING}

Sept. 2020 -- \textbf{Heden}:
  Data Engineer
  \hlmask{} Solutions NV, Brussel
  Verantwoordelijk voor ETL-pipelines ...
  - Tools: Python, Spark, dbt

Jan. 2016 -- Aug. 2020:
  Junior Ontwikkelaar
  InnoSoft BVBA, Gent
  - Back-end Java/Spring Boot; REST APIs
  - Tools: Java, Spring Boot, PostgreSQL

Mrt. 2015 -- Dec. 2015:
  \textbf{Stage} Softwareontwikkelaar
  TechStart SA, Antwerpen

\textbf{OPLEIDING}
2013--2015  Master in Informatica
            Universiteit Gent
2010--2013  Bachelor Toegepaste Informatica
            \textbf{Hogeschool} Gent

\textbf{TALEN}
Nederlands: \textbf{Moedertaal}
Engels: \textbf{Vloeiend} | Frans: Goed

\textbf{CERTIFICATEN}
Nov. 2021: AWS Certified Solutions Architect
           Amazon Web Services
\end{alltt}}
  \tcblower
  \footnotesize\textit{\textbf{Metadata}: Captured at February 2022}
  \end{tcolorbox}
\end{minipage}
\hfill
\begin{minipage}[t]{0.53\textwidth}
  \vspace{0pt}
  {\small\bfseries (b) Structured JSON output (extracted)}\\[1pt]
  \begin{tcolorbox}[
    enhanced, colback=blue!2, colframe=blue!30,
    boxrule=0.5pt, arc=3pt,
    left=5pt, right=4pt, top=5pt, bottom=5pt
  ]
  {\scriptsize\begin{alltt}
\{\jkey{work\_experiences}: [\{
   \jkey{title}: \jval{Data Engineer},
   \jkey{standard\_title}: \jval{Data engineer},\jnote{ESCO-aligned}
   \jkey{company\_name}: \jval{-},\jnote{masked \ensuremath{\to} \texttt{"-"}}
   \jkey{location}: \jval{Brussel},
   \jkey{start\_date}: \jval{Sept. 2020},
   \jkey{end\_date}: \jval{Feb. 2022},\jnote{Heden \ensuremath{\to} timestamp}
   \jkey{description}: \jval{Verantwoordelijk...},
   \jkey{contract\_type}: \jval{employment},
   \jkey{skills\_used}: [\jval{Python}, \jval{Spark}, \jval{dbt}]
 \}, \{
   \jkey{title}: \jval{Junior Ontwikkelaar},
   \jkey{standard\_title}: \jval{Software developer},
   \jkey{company\_name}: \jval{InnoSoft BVBA},
   \jkey{start\_date}: \jval{Jan. 2016},
   \jkey{end\_date}: \jval{Aug. 2020},
   \jkey{contract\_type}: \jval{employment},
   \jkey{skills\_used}: [\jval{Java}, \jval{Spring Boot}, ...]
 \}, \{
   \jkey{title}: \jval{Stage Softwareontwikkelaar},
   \jkey{standard\_title}: \jval{Software developer},
   \jkey{contract\_type}: \jval{internship},\jnote{Stage \ensuremath{\to} internship}
   \jkey{start\_date}: \jval{Mrt. 2015},
   \jkey{end\_date}: \jval{Dec. 2015},
   \jkey{skills\_used}: [\jval{Django}]
 \}],
 \jkey{educations}: [\{
   \jkey{degree\_level}: \jval{Master},
   \jkey{degree\_title}: \jval{Master in Informatica},
   \jkey{institute\_name}: \jval{Universiteit Gent},
   \jkey{start\_date}: \jval{2013}, \jkey{end\_date}: \jval{2015}
 \}, \{
   \jkey{degree\_level}: \jval{Bachelor},\jnote{Hogeschool \ensuremath{\to} Bachelor}
   \jkey{degree\_title}: \jval{Bachelor Toegepaste Informatica},
   \jkey{institute\_name}: \jval{Hogeschool Gent},
   \jkey{start\_date}: \jval{2010}, \jkey{end\_date}: \jval{2013}
 \}],
 \jkey{languages}: [
   \{\jkey{language}: \jval{Nederlands},
    \jkey{proficiency\_level}: \jval{native}\},\jnote{Moedertaal \ensuremath{\to} native}
   \{\jkey{language}: \jval{Engels},
    \jkey{proficiency\_level}: \jval{fluent}\},\jnote{Vloeiend \ensuremath{\to} fluent}
   \{\jkey{language}: \jval{Frans},
    \jkey{proficiency\_level}: \jval{intermediate}\}
 ],
 \jkey{other\_certificates}: [\{
   \jkey{certificate\_name}: \jval{AWS Cert. Solutions Architect},
   \jkey{issuing\_organization}: \jval{Amazon Web Services},
   \jkey{issue\_date}: \jval{Nov. 2021}
 \}]\}
\end{alltt}}
  \end{tcolorbox}
\end{minipage}
\caption{%
  Extraction illustration using a synthetic resume representative of the VDAB corpus
  (Section~\ref{sec:JobHop_v2_dataset}).
  \textbf{(a)}~Pseudonymized input: \colorbox{orange!40}{\ttfamily\tiny<MASK>}~tokens replace
  removed personally identifiable information; section headers and descriptions appear in Dutch.
  Bold terms mark Dutch-specific signals.
  \textbf{(b)}~Structured JSON produced by the extraction prompt, demonstrating five pipeline
  behaviors:
  (\emph{i})~masked fields resolved to~\texttt{"-"};
  (\emph{ii})~present-tense marker \emph{Heden} resolved to the capture timestamp;
  (\emph{iii})~free-text descriptions extracted and preserved (shown truncated);
  (\emph{iv})~original Dutch job titles mapped to ESCO-aligned \texttt{standard\_title} values;
  (\emph{v})~Dutch semantic conventions normalized to schema labels
  (\emph{Stage}~$\to$~\texttt{internship}, \emph{Hogeschool}~$\to$~\texttt{bachelor}, \emph{Moedertaal}~$\to$~\texttt{native}, \emph{Vloeiend}~$\to$~\texttt{fluent}).
}
\label{fig:extraction_example}
\end{figure*}

\section{Inference Infrastructure and Throughput}\label{app:inference}

This section specifies the compute environment shared by all experiments reported in the paper, together with the inference configuration used for the production extraction pipeline.

\subsection*{Hardware}

All experiments were conducted on a single workstation equipped with two NVIDIA H200~NVL GPUs ($140$\,GB HBM3e each;
$280$\,GB combined GPU memory), a dual-socket AMD EPYC~9535 CPU ($2\times64$ cores, $256$ hardware threads),
and $1.5$\,TiB of system RAM\@. CUDA~12.8 was used throughout.

\subsection*{LLM Extraction}

All LLM-based extraction was performed on this machine using vLLM~\cite{kwon2023efficient} for batched inference.
Key inference parameters for the production extraction pipeline:

\begin{center}
\begin{tabular}{ll}
\toprule
\textbf{Parameter} & \textbf{Value} \\
\midrule
Model & \texttt{openai/gpt-oss-120b} \\
Precision & bfloat16 \\
Tensor parallelism & 2 (one shard per GPU) \\
Prefix caching & Enabled \\
Decoding & Greedy ($T{=}0$) \\
Repetition penalty & 1.1 (initial), 1.15 (retry) \\
Max generation tokens & 24,000 (initial), 32,000 (retry) \\
Reasoning effort & \textsc{high} (both passes) \\
Batch size & 200 resumes per batch \\
GPU memory utilization & 0.90 \\
\bottomrule
\end{tabular}
\end{center}

The retry mechanism targets samples that fail JSON validation after the initial pass. Of approximately $400{,}000$ input
resumes, ${\sim}50{,}500$ ($12.6\%$) failed the initial pass; all were recovered through the retry loop with an extended
generation budget and increased repetition penalty.
The final corpus therefore has a $100\%$ JSON parse rate.

For the evaluation experiments (Table~\ref{tab:ie_model_comparison}), GPT-OSS-120B was evaluated across $10$ independent
extraction runs on the $200$-sample benchmark.
Each run used the same prompt and inference configuration, producing the $95\%$ confidence intervals reported in the main
text.

\section{Data Cleaning Pipeline}\label{app:cleaning}

This section details the five-stage cleaning pipeline applied to the JobHop~v2 extraction outputs
(Section~\ref{sec:JobHop_v2_dataset}). The five stages are applied sequentially, each operating on the output of the previous one.

\subsection*{Stage 1: Parsing-Artifact Correction}

Dates containing extraction artifacts are identified via regular expression patterns targeting:
\begin{itemize}
    \item Residual \texttt{<MASK>} tokens within date fields
    \item Future-year placeholders (years $> 2025$)
    \item Malformed date strings with dangling hyphens or slashes
\end{itemize}
Affected dates are either corrected to the extractable year component or set to missing.

\subsection*{Stage 2: Invalid Range Correction}

Entries where the parsed start date is later than the end date are flagged. Where the discrepancy is small ($\leq 1$
quarter), the dates are swapped; entries with implausible durations ($> 40$ years) are removed.

\subsection*{Stage 3: Exact Deduplication}

Entries identical across all core fields (title, company, start date, end date) within the same resume are collapsed to a
single entry. This addresses extraction artifacts where multi-column resume layouts cause the same position to be
extracted twice.

\subsection*{Stage 4: Quality Filtering}

Resumes are removed if they contain:
\begin{itemize}
    \item More than $20$ work-experience entries (likely parsing errors on non-resume documents)
    \item Zero meaningful entries across all entity groups after earlier cleaning stages
\end{itemize}

\subsection*{Stage 5: Consecutive-Job Merging}

Adjacent entries within the same resume are merged if they share:
\begin{itemize}
    \item The same normalized job title
    \item The same normalized company name
    \item Temporally contiguous or overlapping date ranges (gap $\leq 1$ quarter)
\end{itemize}
Descriptions are concatenated with a separator. This stage consolidates approximately $19{,}432$ entries that were split
during extraction due to multi-line or multi-column resume formatting.

\section{Normalization Pipeline}\label{app:normalization}

After cleaning (Appendix~\ref{app:cleaning}), the extracted fields are normalized into controlled representations suited to cross-resume analysis and modeling. Normalization comprises three independent components, described in turn below: assigning ESCO occupation codes to job entries, converting dates to quarter-year granularity, and mapping education entries to a common degree-level taxonomy.

\subsection*{Occupation Code Assignment}

Individual job titles are highly variable, so cross-resume analysis requires mapping them to a controlled
vocabulary~\cite{llm4jobs}. We adopt the ESCO taxonomy (v1.1.2)~\cite{le2014esco} as the target vocabulary;
it is organized hierarchically with up to $3{,}007$ leaf-node occupations at levels five through seven.
Occupation-code assignment uses a commercial classifier provided by
Nobl.ai,\footnote{\href{https://nobl.ai/}{https://nobl.ai/}} applied via a two-step hybrid policy that exploits the richer
JobHop~v2 extraction schema:

\textbf{Primary assignment.}
The concatenation of the extracted \texttt{title} and
\texttt{description} fields is passed to the classifier,
and the top-1 prediction is accepted if its confidence exceeds $\tau_1 = 0.45$.

\textbf{Rescue assignment.}
For entries below $\tau_1$ in primary assignment, the model-generated \texttt{standard\_title} field is submitted to the
same classifier, and the top-1 prediction is accepted only if confidence exceeds a stricter threshold $\tau_2 = 0.85$.
Entries remaining unassigned after both steps, or with uninformative descriptions, are labeled \texttt{unknown}.
The asymmetric thresholds are deliberate: primary assignment maximizes coverage at moderate confidence,
whereas rescue assignment acts as a high-precision fallback activated only when the standardized title provides a strong
signal.

\subsection*{Temporal Normalization}

All dates are converted to quarter-year format (\texttt{Q\# YYYY}) through the following procedure:

\begin{enumerate}
    \item \textbf{Month name normalization.} A multilingual mapping converts Dutch (``januari'', ``februari'', ``maart'', ...), French (``janvier'', ``f\'{e}vrier'', ``mars'', ...), and English month names to a canonical numeric representation.
    \item \textbf{Pattern extraction.} Regular expressions extract year and optional month from heterogeneous date formats including: ``MM-YYYY'', ``YYYY-MM'', ``Month YYYY'', ``MM/YYYY'', year-only, and abbreviated forms (``jan.\ 2015'', ``sept./d\'{e}c.\ 2004'').
    \item \textbf{Quarter mapping.} Months 1--3 $\to$ Q1, months 4--6 $\to$ Q2, months 7--9 $\to$ Q3, months 10--12 $\to$ Q4. Month-only dates default to Q1.
    \item \textbf{Validity checks.} Years outside the range $[1950, 2025]$ are rejected; the resulting quarter string is validated against the pattern \texttt{Q[1-4] [0-9]\{4\}}.
\end{enumerate}

\subsection*{Education Level Normalization}

Education entries are mapped to a five-level taxonomy using the following mapping:

\begin{center}
\small
\begin{tabular}{ll}
\toprule
\textbf{Input terms (multilingual)} & \textbf{Normalized level} \\
\midrule
Doctoraat, PhD, Doctorate & PhD \\
Master, Licentiate, Lic., Ma. & Master \\
Bachelor, Hogeschool, Professional Bachelor, Ba. & Bachelor \\
Secondary, Middelbaar, Secondaire, ASO, TSO, BSO & Secondary \\
Primary, Lager onderwijs, Primaire, and other degrees & None \\
\bottomrule
\end{tabular}
\end{center}

When the \texttt{degree\_level} field is missing or uninformative, keyword matching is applied to the
\texttt{degree\_title} field.
This fallback recovers degree-level information for approximately $3{,}194$ entries.
The highest attained degree level per resume is computed by taking the maximum across all education entries,
and is used as the degree input for the prediction model (Section~\ref{sec:pred_film}).

\section{JobHop~v2: Dataset Summary}\label{app:dataset_summary}

Compared with prior public benchmarks, JobHop~v2 is the first large-scale, multilingual dataset constructed through full
end-to-end extraction from real unstructured resumes. Decorte et al.~\cite{decorte2023career} extracted $2{,}164$ career
histories from Kaggle resumes using GPT-3.5 at considerably smaller scale and without temporal metadata,
education annotations, or multilingual coverage. OpenResume~\cite{openresume} likewise performs extraction from Kaggle
resumes but releases only $301$ real (English) trajectories supplemented with synthetic variants.
Karrierewege~\cite{karrierewege} provides $568{,}888$ career paths from the German Federal Employment Agency,
but the underlying data consists of Berufenet occupational codes rather than raw resume text;
free-text descriptions were synthesized rather than extracted. JobHop~v2 addresses a more complex and realistic extraction
scenario: pseudonymized, multilingual, unstructured documents requiring simultaneous extraction of multiple entity types
with rich attribute schemas.

\begin{figure*}[t]
  \centering
  \begin{subfigure}[b]{0.58\textwidth}
    \centering
    \includegraphics[width=\linewidth]{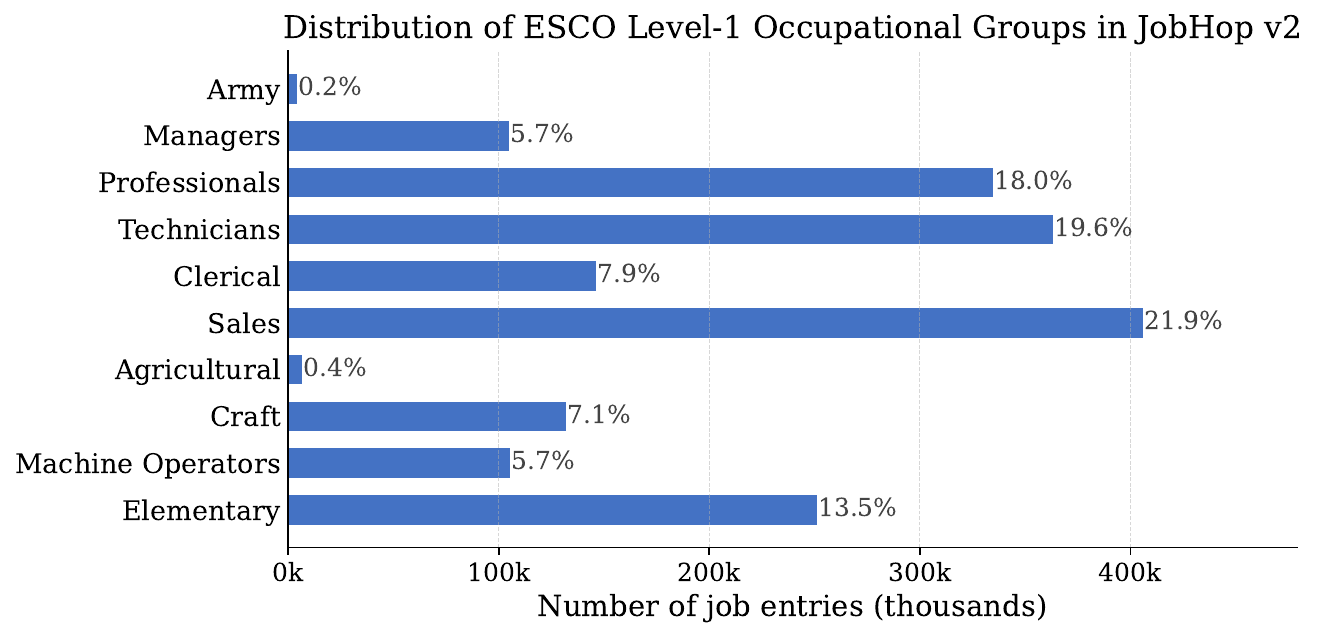}
    \caption{Distribution of ESCO level-1 (ISCO major group) occupational categories.}
    \label{fig:esco_distribution}
  \end{subfigure}
  \hfill
  \begin{subfigure}[b]{0.40\textwidth}
    \centering
    \includegraphics[width=\linewidth]{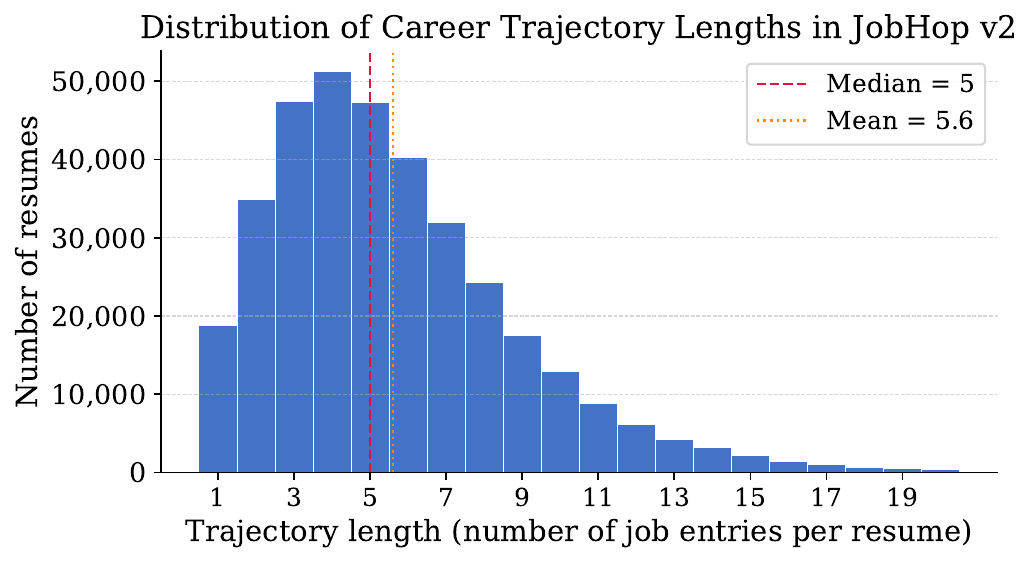}
    \caption{Distribution of career trajectory lengths (job entries per resume).}
    \label{fig:trajectory_lengths}
  \end{subfigure}
  \caption{Occupational diversity and career history lengths in JobHop~v2.
           Professionals (ISCO~2), service~\&~sales (ISCO~5), and Technicians~\&~Associate Professionals (ISCO~3)
           account for the majority of job entries, reflecting the white-collar
           focus of the underlying resume corpus.
           Trajectory lengths range from 1 to over 20 jobs, with a median of~5.}
  \label{fig:dataset_distributions}
\end{figure*}

Figure~\ref{fig:dataset_distributions} characterises the dataset along two complementary dimensions.
Figure~\ref{fig:esco_distribution} shows the distribution of job entries across ESCO level-1 occupational groups (ISCO-08
major groups). Service~\&~Sales Workers (ISCO~5), Professionals (ISCO~2), and Technicians~\&~Associate Professionals
(ISCO~3) account for the largest share of entries, reflecting the white-collar composition of the underlying resume
corpus. Elementary occupations (ISCO~9) are the next most frequent group; primary-sector occupations (ISCO~6,
Skilled Agricultural) are rare, reflecting the labour-market profile of the source population.
Figure~\ref{fig:trajectory_lengths} reports the distribution of career trajectory lengths,
measured as the number of job entries per resume after cleaning. The distribution is right-skewed:
most individuals have between two and six recorded positions, but a non-trivial fraction of resumes contain ten or more
entries, providing rich sequential context for the prediction models in Section~\ref{sec:prediction}.
Together, the two panels confirm that JobHop~v2 captures a broad occupational spectrum and a realistic range of
career-history depths.

\section{Extraction Evaluation Protocol}\label{app:evaluation_protocol}

This section provides the complete parameterization of the revised evaluation protocol used for the extraction
evaluation reported in Appendix~\ref{app:extraction_eval}. The protocol normalizes each field, scores it with a string-similarity or date-specific function, weights fields by importance, and composes these components into a single sample-level score, as detailed in the four parts below.

\subsection*{String Similarity Scoring}

The string similarity function operates on normalized text (lowercased, whitespace-collapsed,
trailing punctuation removed, common abbreviations unified). After normalization:

{\small
\begin{enumerate}
    \item \textbf{Exact match:} score $= 100$.
    \item \textbf{Both fields empty/missing:} score $= 100$.
    \item \textbf{One field meaningful, other missing:} score $= 10$ (penalty).
    \item \textbf{Substring containment:}
If one normalized string is contained in the other, the score depends on the length ratio $r$:
    
    \begin{center}
        $r > 0.7$: score $= 90$ \qquad\qquad $0.5 < r \leq 0.7$: score $= 75$ \qquad\qquad $r \leq 0.5$: score $= 50$
    \end{center}
    \item \textbf{Fuzzy matching} (Levenshtein ratio $s$):
    \begin{center}
        $s \geq 90$: score $= s$ (no penalty) \qquad\qquad $s < 60$: score $= \max(0,\; s - 20)$ 
        
        $60 \leq s < 70{:}\ s - 15$ \quad\quad\quad $70 \leq s < 80{:}\ s - 10$ \quad\quad\quad $80 \leq s < 90{:}\ s - 5$
    \end{center}
\end{enumerate}}

\subsection*{Date Scoring}

Dates are parsed flexibly, supporting English, Dutch, and French month names. Present/ongoing markers (``Present'',
``heden'', ``nu'') are mapped to a reference date.

{\small
\begin{itemize}
    \item Both unparseable (text present but not interpretable): score $= 50$.
    \item One parseable, one not: score $= 20$.
    \item Year and month match: score $= 85$.
    \item Year-only match: score $= 60$.
    \item Month match with close year ($|\Delta\text{year}| \leq 1$): score $= 40$.
    \item No match: score $= 15$.
    \item Both dates missing: score $= 30$.
    \item One date missing: score $= 30$.
\end{itemize}}

\subsection*{Field Importance Weights}

{\small
\begin{itemize}
  \item Title, Company/Institute Name: $1.5\times$
  \item Start Date, End Date: $1.3\times$
  \item Description: $1.2\times$
  \item Place/Location, Type: $1.0\times$
  \item Degree Level: $0.9\times$
  \item Other fields: $1.0\times$ (default)
\end{itemize}}


\subsection*{Sample-Level Score Composition}

The overall score per sample is computed as:
\begin{equation*}
    S = P_{\text{count}} \cdot \bigl(0.6 \cdot \bar{M} + 0.2 \cdot R_{\text{entry}} + 0.2 \cdot P_{\text{entry}}\bigr),
\end{equation*}
where $\bar{M}$ is the mean matched-entry weighted similarity (computed via the Hungarian algorithm),
$R_{\text{entry}}$ and $P_{\text{entry}}$ are entry-level recall and precision, and $P_{\text{count}}$ is an entry-count
penalty:
\begin{equation*}
    P_{\text{count}} = \max\bigl(0.5,\; 1.0 - 0.3 \cdot |n_{\text{gt}} - n_{\text{pred}}| / \max(n_{\text{gt}}, n_{\text{pred}})\bigr).
\end{equation*}

Special cases: both ground truth and prediction empty $\to$ score $= 100$; ground truth empty but prediction non-empty
$\to$ score $= 10$; prediction empty but ground truth non-empty $\to$ score $= 5$.

\section{Extraction Evaluation: Datasets and Results}\label{app:extraction_eval}

Reliable evaluation of LLM-based extraction requires both a scoring protocol that faithfully reflects quality for
downstream use and reference annotations that minimize labeling noise as a confound. Section~\ref{app:evaluation_protocol}
describes the revised evaluation protocol; this section reports the labeled reference sets and model comparison results.

\subsection{Labeled Datasets}\label{app:label_baselines}

Extraction quality is evaluated on 200 benchmark resumes drawn from the set annotated in the original JobHop
study~\cite{jobhop-v1}. In that work, the authors and external annotators produced hand annotations over approximately 60
cumulative hours, yielding the primary ground-truth reference. During development of the revised protocol,
we observed that these annotations contain occasional inconsistencies (divergent date-field granularity,
differing conventions for ambiguous section boundaries, and subjective judgment calls varying across annotators)
that can systematically bias accuracy estimates.

To obtain a more complete evaluation picture, we construct two additional labeled datasets:

\begin{enumerate}
    \item \textbf{Hand annotations.}
The original multi-annotator labels, reflecting genuine human judgment but exhibiting inter-annotator variability.

    \item \textbf{Edited labels.}
A systematic revision by an AI agent (Claude Sonnet) instructed to resolve inconsistencies and enforce uniform
conventions while preserving correct annotations; this baseline reduces labeling noise while retaining the structure of
the original hand annotations.

    \item \textbf{Agent labels.}
An independent annotation set generated by the same AI agent \emph{without access to the hand or edited annotations};
by eliminating dependence on the original human labels, this baseline provides an internally consistent reference
unanchored to the conventions or errors of the original annotation process.
\end{enumerate}

These three baselines form a progression in annotation consistency: from multi-annotator hand labels (highest
variability), through edited labels (reduced noise), to agent labels (most internally consistent).
Evaluating against all three disentangles genuine extraction errors from annotation noise and assesses whether comparative
conclusions are robust to the choice of reference.

\subsection{Results}\label{app:extraction_results}

Four extraction systems are compared under the revised protocol: GPT-OSS-120B with \textsc{high} and
\textsc{medium} reasoning, GPT-OSS-20B,
Llama-3.3-70B~\cite{llama3}, and Gemma-2~\cite{gemma2}. GPT-OSS-120B is evaluated across 10 independent extraction runs;
95\% confidence intervals are computed via the $t$-distribution with 9 degrees of freedom.

\begin{table}[t]
\centering
\caption{Extraction accuracy breakdown (\%) across all three labeled datasets.}
\label{tab:ie_model_comparison}
\resizebox{\columnwidth}{!}{%
\begin{threeparttable}
\begin{tabular}{lcccccccccccc}
\toprule
& \multicolumn{4}{c}{\textbf{Hand Ann.}} & \multicolumn{4}{c}{\textbf{Edited}} & \multicolumn{4}{c}{\textbf{Agent}} \\
\cmidrule(lr){2-5} \cmidrule(lr){6-9} \cmidrule(lr){10-13}
\textbf{Model} & \textbf{Sim. (60\%)} & \textbf{Rec. (20\%)} & \textbf{Prec. (20\%)} & \textbf{Final avg} & \textbf{Sim. (60\%)} & \textbf{Rec. (20\%)} & \textbf{Prec. (20\%)} & \textbf{Final avg} & \textbf{Sim. (60\%)} & \textbf{Rec. (20\%)} & \textbf{Prec. (20\%)} & \textbf{Final avg} \\
\midrule
Gemma-2                        & 79.4            & 96.2            & 92.6            & 81.6            & 80.3            & 95.6            & 97.1            & 84.8            & 83.1            & 96.6            & 95.7            & 86.0            \\
Llama-3.3-70B                  & 79.1            & \textbf{97.2}   & 93.9            & 82.5            & 79.0            & \textbf{96.5}   & 97.7            & 84.8            & 83.4            & \textbf{97.8}   & 95.7            & 86.6            \\
GPT-OSS-20B                    & 78.9            & 94.9            & 94.6            & 82.0            & 78.2            & 93.6            & 97.8            & 83.2            & 82.7            & 95.1            & 96.3            & 85.5            \\
GPT-OSS-120B (\textsc{medium}) & $80.3 \pm 0.4$ & $94.7 \pm 0.3$ & $\mathbf{95.7 \pm 0.4}$ & $83.3 \pm 0.1$ & $79.3 \pm 0.5$ & $93.5 \pm 0.5$ & $\mathbf{98.9 \pm 0.1}$ & $84.4 \pm 0.5$ & $84.2 \pm 0.5$ & $94.9 \pm 0.4$ & $\mathbf{97.0 \pm 0.3}$ & $86.7 \pm 0.4$ \\
GPT-OSS-120B (\textsc{high})   & $\mathbf{81.9 \pm 0.2}$ & $96.4 \pm 0.3$ & $93.9 \pm 0.1$ & $\mathbf{83.9 \pm 0.2}$ & $\mathbf{81.7 \pm 0.2}$ & $96.0 \pm 0.2$ & $97.6 \pm 0.2$ & $\mathbf{86.2 \pm 0.2}$ & $\mathbf{86.4 \pm 0.2}$ & $97.0 \pm 0.2$ & $95.5 \pm 0.2$ & $\mathbf{88.0 \pm 0.2}$ \\
\midrule
Label agreement\tnote{a}       & 84.4 & 96.6 & 95.8 & 86.6            & 86.2 & 97.1 & 96.7 & 88.5            & 87.4 & 97.0 & 96.6 & 89.1            \\
\bottomrule
\end{tabular}
\begin{tablenotes}[flushleft]
\scriptsize
\item Similarity, Recall, and Precision are sample-level component means under the revised evaluation protocol; Final avg is the weighted score after entry-count penalty.
\item GPT-OSS-120B values are mean $\pm$ 95\% CI from 10 independent runs. Bold = best model result per column (label agreement excluded).
\item[a] Agreement baseline computed by symmetric pairwise evaluation between label sets (A$\leftrightarrow$B, averaged across both directions), then averaged across the two pairings involving each reference set; reported for Sim., Rec., Prec., and Final avg.
\end{tablenotes}
\end{threeparttable}%
}
\end{table}

Table~\ref{tab:ie_model_comparison} presents the results. GPT-OSS-120B (\textsc{high}) achieves the highest accuracy
across all three baselines ($83.9\%$, $86.2\%$, $88.0\%$), with tight confidence intervals ($\pm 0.2$~pp)
confirming run-to-run reproducibility.
\textsc{High} reasoning consistently outperforms \textsc{medium} by $0.6$--$1.8$~pp, justifying the additional inference cost.
Llama-3.3-70B performs competitively, approaching GPT-OSS-120B (\textsc{medium}) on agent labels.
Gemma-2 achieves comparable accuracy on edited and agent labels despite a lower JSON parse rate (${\approx}86\%$ vs.\
${\geq}98.5\%$), indicating that its errors concentrate in format compliance rather than extraction quality.

The label-agreement row provides an interpretive ceiling: the gap between the best extractor ($83.9\%$ on hand
annotations) and the inter-baseline agreement ceiling ($86.6\%$) is approximately $2.7$~pp,
indicating that much of the residual error reflects genuine annotation ambiguity rather than systematic model failure.
The monotonic increase in reported accuracy from hand annotations through edited labels to agent labels,
consistent across all models, corroborates the annotation-consistency progression described above:
labeling noise in the original hand annotations depresses reported accuracy.

\section{Embedding Model Exploration}\label{app:embedding_exploration}

This section reports the systematic embedding model exploration referenced in Section~\ref{sec:representation}.
All experiments were conducted on a $20\%$ evaluation subset of the Karrierewege dataset using the validation split.
The search explored four base encoders, multiple loss functions, and variations of the TSDAE domain adaptation stage,
yielding $13$ completed configurations.

\subsection*{Experimental Configurations}

\begin{table}[h!]
\centering
\caption{Summary of the $13$ completed embedding training configurations explored during model selection. All evaluated on a $20\%$ subset of the Karrierewege validation set. ``Career FT'' denotes supervised contrastive fine-tuning on career trajectory pairs. Configurations marked with $\star$ were additionally tested with TSDAE pre-adaptation.}
\label{tab:embedding_exploration}
\small
\begin{tabular}{clllccc}
\toprule
\textbf{\#} & \textbf{Base Model} & \textbf{Loss} & \textbf{TSDAE} & \textbf{MRR} & \textbf{R@5} & \textbf{R@10} \\
\midrule
1 & \texttt{multilingual-e5-base} & none (zero-shot) & No & 25.8 & 37.5 & 41.9 \\
2 & \texttt{multilingual-e5-base} & CachedMNRL & No & 28.7 & 31.7 & 35.3 \\
3 & \texttt{all-mpnet-base-v2} & CachedMNRL & Yes$\star$ & 27.9 & 30.6 & 34.0 \\
4 & \texttt{multilingual-e5-base} & CachedMNRL & Yes$\star$ & 28.7 & 31.7 & 35.4 \\
5 & \texttt{multilingual-e5-base} & CachedMNRL & Yes (aug.) & 28.5 & 31.3 & 34.9 \\
6 & \texttt{multilingual-e5-base} & CachedMNRL & Yes (5 ep.) & 28.6 & 31.6 & 35.3 \\
7 & \texttt{multilingual-e5-base} & CachedMNRL & Yes (del.~0.3) & 28.6 & 31.4 & 35.1 \\
8 & \texttt{multilingual-e5-base} & CachedMNRL + HN & No & 29.6 & 33.4 & 38.7 \\
9 & \texttt{multilingual-e5-base} & CachedMNRL + HN & Yes$\star$ & 29.6 & 33.6 & 38.5 \\
10 & \texttt{multilingual-e5-large} & CachedMNRL & No & 28.1 & 30.3 & 33.4 \\
11 & \texttt{bge-large-en-v1.5} & CachedMNRL & No & 28.3 & 29.9 & 33.1 \\
12 & \texttt{multilingual-e5-base} & MatryoshkaLoss & No & 28.6 & 31.3 & 35.0 \\
13 & \texttt{multilingual-e5-base} & GISTEmbedLoss & No & 30.3 & 36.3 & 40.9 \\
\bottomrule
\end{tabular}
\end{table}

\subsection*{Key Findings}

\begin{enumerate}
    \item \textbf{Base model selection.}
Among the five base encoders,
\texttt{multilingual-e5-base} (768-dim)
achieved the highest MRR among base-scale models. The 1024-dimensional variants (\texttt{multili ngual-e5-large},
\texttt{bge-large-en-v1.5}) showed no consistent improvement despite approximately $2\times$ the parameter count and
substantially longer training times. The multilingual pretraining of E5 is particularly relevant for the
Dutch/French/English composition of the JobHop corpus.

    \item \textbf{Loss function comparison.}
GISTEmbedLoss (Exp.~13) achieved the best overall recall among non-TSDAE configurations.
The guided in-sample negative selection filters false negatives arising from ESCO taxonomy structure (semantically similar
occupations within the same group), yielding cleaner gradients than standard MNRL. MatryoshkaLoss (Exp.~12)
enables flexible dimensionality reduction but did not improve full-dimensional performance.

    \item \textbf{TSDAE domain adaptation.}
TSDAE pre-adaptation before supervised fine-tuning consistently improved recall metrics. The default configuration ($3$
epochs, $0.6$ deletion ratio, ESCO-only text corpus) proved most effective.
Augmenting the TSDAE corpus with career trajectory text (Exp.~5) provided marginal additional benefit;
longer TSDAE training ($5$ epochs, Exp.~6) and lower deletion ratios ($0.3$, Exp.~7) did not yield improvements over the
default.

    \item \textbf{Hard negative mining.}
Using ESCO taxonomy structure to construct hard negatives from the same ISCO group (Exps.~8--9)
did not consistently improve over standard in-batch negatives, likely because the ESCO hierarchy is relatively flat (up to
$7$ levels) and GISTEmbedLoss already addresses false-negative contamination via its guide model.

    \item \textbf{Selected configuration.}
Based on these results, the final ROUTE pipeline uses
\texttt{multilingual -e5-base} with TSDAE domain
adaptation followed by GISTEmbedLoss contrastive fine-tuning (combining the insights of Exps.~4 and~13),
corresponding to the two-stage method described in Section~\ref{sec:representation}.
\end{enumerate}

\section{Embedding Fine-Tuning Hyperparameters}\label{app:embedding_hyperparams}

Table~\ref{tab:embedding_hyperparams_combined} lists the complete hyperparameter configuration for each stage of the
embedding fine-tuning pipeline and the single-stage baseline described in Section~\ref{sec:representation}.

\begin{table}[h!]
\centering
\caption{Hyperparameters for the two-stage embedding fine-tuning pipeline and the single-stage baseline. Stage~2 initialises from the checkpoint produced by Stage~1. Dashes indicate parameters not applicable to that stage.}
\label{tab:embedding_hyperparams_combined}
\small
\begin{tabular}{llll}
\toprule
\textbf{Hyperparameter} & \textbf{Stage 1 (TSDAE)} & \textbf{Stage 2 (Contrastive)} & \textbf{Base} \\
\midrule
Base model        & \texttt{multilingual-e5-base} & (from Stage~1)               & \texttt{all-mpnet-base-v2} \\
Loss              & DenoisingAutoEncoderLoss      & GISTEmbedLoss                & MNRL \\
Guide model       & —                             & \texttt{all-MiniLM-L6-v2}    & — \\
Deletion ratio    & 0.6                           & —                            & — \\
Corpus            & ESCO + career segments        & —                            & — \\
Epochs            & 3                             & 2                            & 2 \\
Batch size        & 32                            & 64                           & 320 \\
Learning rate     & $3\times10^{-5}$              & $2\times10^{-5}$             & $2\times10^{-5}$ \\
Weight decay      & 0.01                          & 0.01                         & 0 \\
Warmup            & 10\% of steps                 & 5\% of steps                 & 5\% of steps \\
LR schedule       & —                             & —                            & Linear \\
Precision         & FP16                          & FP16                         & — \\
Evaluation        & Every $\lfloor N_\text{step}/4\rfloor$ & Every 10\% of epoch & — \\
Best checkpoint   & Lowest val.\ loss             & Highest val.\ R@10           & — \\
\bottomrule
\end{tabular}
\end{table}

\section{STEP: Architecture and Hyperparameters}\label{app:model_architecture}

This section provides implementation details for STEP (Section~\ref{sec:pred_proposed}). We report the parameter budget, the full set of training hyperparameters, and the weight-initialization scheme.

\subsection*{Parameter Count}

STEP has approximately \textbf{221,842 trainable parameters}, distributed as follows:

\begin{center}
\begin{tabular}{lr}
\toprule
\textbf{Component} & \textbf{Parameters} \\
\midrule
Time-Decay GRU Cell (input: 768, hidden: 64) & 160,192 \\
\quad \textit{of which: } \texttt{log\_lambda} (learnable decay rates) & 64 \\
Layer Normalization (64) & 128 \\
FiLM Conditioning & 9,488 \\
\quad Degree embedding ($5 \times 16$) & 80 \\
\quad Condition network ($16 \to 64 \to 128$) & 9,408 \\
Attention Pooling ($64 \to 32 \to 1$) & 2,113 \\
Output Projection ($64 \to 768$) & 49,920 \\
Learnable Temperature & 1 \\
\midrule
\textbf{Total} & \textbf{221,842} \\
\bottomrule
\end{tabular}
\end{center}

For comparison, the BiLSTM baseline (hidden dimension $16$, $32$ bidirectional) has approximately $102{,}400$ parameters,
and the MLP baseline (hidden dimension $512$) has approximately $787{,}968$ parameters.
STEP is lightweight relative to the MLP, adding negligible overhead above the embedding computation.

\subsection*{Hyperparameters}

\paragraph{Architecture dimensions.}
The GRU hidden dimension is $64$. The FiLM conditioning branch embeds degree level into a $16$-dimensional space over $5$
classes and projects through a two-layer network before producing scale and shift parameters for the hidden state.

\paragraph{Regularization.}
Dropout ($p = 0.1$) is applied after layer normalization. Label smoothing with $\epsilon = 0.1$ is used during training to
prevent overconfident predictions on the large ESCO label space. Gradients are clipped to a maximum norm of $1.0$.

\paragraph{Optimization and schedule.}
The model is trained with AdamW ($\beta_1{=}0.9$, $\beta_2{=}0.999$), a learning rate of $10^{-3}$,
and weight decay $10^{-4}$. The learning rate follows a cosine annealing schedule over the training budget.

\paragraph{Clamping ranges.}
The learnable temperature $\tau$ is initialized at $0.07$ and clamped to $[10^{-3},\, 1.0]$ throughout training.
Inter-job time intervals fed to the time-decay GRU are clamped to $[0.1,\, 20.0]$ years to suppress the influence of
data-quality outliers.

\paragraph{Training procedure.}
Models are trained for up to $10$ epochs ($30$ for the Decorte dataset, which is substantially smaller)
with a batch size of $64$. Early stopping is applied on the basis of validation R@10, and the best checkpoint is retained
for evaluation. All stochastic results are averaged over $5$ independent runs with different random seeds.

\subsection*{Weight Initialization}

All linear layers use Xavier uniform initialization; all biases are initialized to zero. The decay rate parameters
$\boldsymbol{\ell}$ (parameterizing $\boldsymbol{\lambda} = \text{softplus}(\boldsymbol{\ell})$)
are initialized at zero, producing $\boldsymbol{\lambda} \approx 0.693$ at the start of training (approximately no decay
for $\delta_t = 1$ year). The temperature parameter is initialized as $\log(\tau_0) = \log(0.07) \approx -2.66$.

\section{Additional Career Path Recommendation Results}\label{app:additional_results}

Table~\ref{tab:cpr_additional} reports career path recommendation performance on three additional derivative datasets not
included in the main text: Decorte-ESCO (the ESCO-description variant of Decorte), Karrierewege-Occ (the
occupation-title-only variant of Karrierewege), and Karrierewege-CP (the career-path variant of Karrierewege).
These datasets derive from the same underlying sources as Decorte and Karrierewege, respectively,
but use different text representations or subsequence strategies.
None include temporal or degree annotations; all models therefore run in reduced mode.

\begin{table*}[h!]
\centering
\caption{Career path recommendation on three additional derivative datasets. All values are percentages. Stochastic models report mean $\pm$ std over 5 runs. Best per dataset and metric in \textbf{bold}.}
\label{tab:cpr_additional}
\resizebox{\textwidth}{!}{%
\begin{tabular}{l ccc ccc ccc}
\toprule
& \multicolumn{3}{c}{\textbf{Decorte-ESCO}} & \multicolumn{3}{c}{\textbf{Karrierewege-Occ}} & \multicolumn{3}{c}{\textbf{Karrierewege-CP}} \\
\cmidrule(lr){2-4} \cmidrule(lr){5-7} \cmidrule(lr){8-10}
\textbf{Model} & MRR & R@5 & R@10 & MRR & R@5 & R@10 & MRR & R@5 & R@10 \\
\midrule
Markov                    & 17.7 & 23.4 & 26.0 & 33.7 & 44.4 & 53.2 & 17.3 & 26.9 & 37.4 \\
$\text{ST}_\text{Base}$   & 23.9 & 34.3 & 41.1 & 27.7 & 36.4 & 41.2 & 29.8 & 38.1 & 43.2 \\
$\text{ST}_\text{ROUTE}$  & 24.5 & 34.0 & 40.9 & 26.4 & 34.9 & 42.3 & 26.6 & 37.7 & 44.6 \\
Linear                    & 22.8 & 30.6 & 36.6 & 34.8 & 43.8 & 51.1 & 35.9 & 45.6 & 53.7 \\
MLP                       & $25.8_{\pm 0.6}$ & $36.3_{\pm 0.8}$ & $42.4_{\pm 0.6}$ & $42.3_{\pm 0.1}$ & $56.3_{\pm 0.2}$ & $66.8_{\pm 0.1}$ & $43.2_{\pm 0.1}$ & $58.3_{\pm 0.1}$ & $68.4_{\pm 0.1}$ \\
BiLSTM                    & $22.6_{\pm 0.3}$ & $31.9_{\pm 1.0}$ & $40.5_{\pm 0.5}$ & $44.0$ & $58.2_{\pm 0.1}$ & $68.3_{\pm 0.1}$ & $43.4_{\pm 0.1}$ & $58.4_{\pm 0.1}$ & $68.7_{\pm 0.1}$ \\
\midrule
\textbf{STEP}             & $\mathbf{26.5}_{\pm 0.2}$ & $\mathbf{36.5}_{\pm 0.3}$ & $\mathbf{44.9}_{\pm 0.7}$ & $\mathbf{45.0}_{\pm 0.1}$ & $\mathbf{59.4}$ & $\mathbf{69.6}_{\pm 0.1}$ & $\mathbf{44.6}_{\pm 0.1}$ & $\mathbf{60.1}_{\pm 0.1}$ & $\mathbf{70.4}_{\pm 0.1}$ \\
\bottomrule
\end{tabular}%
}
\end{table*}

These derivative datasets serve as robustness checks: Decorte-ESCO tests whether performance generalizes when inputs use
ESCO taxonomy descriptions rather than free-text job titles, while the Karrierewege variants test sensitivity to the text
representation (occupation titles vs.\ career-path concatenation vs.\ synthesized descriptions).
Across all three, STEP attains the best result on every metric, including Decorte-ESCO R@10 of $44.9\%$ (vs.\ MLP
$42.4\%$, BiLSTM $40.5\%$), Karrierewege-Occ R@10 of $69.6\%$ (vs.\ BiLSTM $68.3\%$, MLP $66.8\%$),
and Karrierewege-CP R@10 of $70.4\%$ (vs.\ BiLSTM $68.7\%$, MLP $68.4\%$). 

\end{document}